\documentclass[11pt]{article}

\usepackage[final]{acl}

\usepackage{times}
\usepackage{latexsym}

\usepackage[T1]{fontenc}
\usepackage[utf8]{inputenc}

\usepackage{microtype}

\usepackage{inconsolata}

\usepackage{graphicx}

\usepackage{makecell}
\usepackage{tabularx}
\usepackage{booktabs}
\usepackage{makecell}
\usepackage{multicol}
\usepackage{multirow}
\usepackage{amsmath}
\usepackage{xcolor}
\usepackage{float}  
\usepackage{todonotes}

\usepackage[most]{tcolorbox}
\definecolor{frame}{HTML}{ccefff}    % setting frame color to be used
\definecolor{back}{HTML}{FAFEFF}     % setting sub color to be used

\definecolor{t_frame}{HTML}{ccefff}    % setting frame color to be used
\definecolor{t_back}{HTML}{ccefff}     % setting sub color to be used
\definecolor{f_blue}{HTML}{6A93CE}
\definecolor{f_yellow}{HTML}{EBB444}
\definecolor{f_red}{HTML}{EA6B66}

\title{From Scoring to Explanations: Evaluating SHAP and LLM Rationales\\for Rubric-based Teaching Quality Assessment}

\author{
  \textbf{Ivo Bueno\textsuperscript{1,2}}~
  \textbf{Babette Bühler\textsuperscript{1, 2}}~
  \textbf{Philipp Stark\textsuperscript{3}}~
  \textbf{Tim Fütterer\textsuperscript{4}}
\\
  \textbf{Ulrich Trautwein\textsuperscript{4}}~
  \textbf{Dorottya Demszky\textsuperscript{5}}~
  \textbf{Heather Hill\textsuperscript{6}}~
  \textbf{Enkelejda Kasneci\textsuperscript{1,2}}
\\
\\
  \textsuperscript{1}Technical University of Munich~
  \textsuperscript{2}Munich Center for Machine Learning (MCML)
\\
  \textsuperscript{3}Lund University~ 
  \textsuperscript{4}University of Tübingen~
  \textsuperscript{5}Stanford Graduate School of Education
\\
  \textsuperscript{6}Harvard Graduate School of Education
\\
  \small{
    \textbf{Correspondence:} \href{mailto:ivo.bueno@tum.de}{ivo.bueno@tum.de}
  }
}

\begin{document}
\maketitle
\begin{abstract}
Automated scoring models are increasingly used to assign rubric-based quality ratings to complex language performances, including classroom transcripts, yet they typically provide little insight into why a particular score is produced. We propose a general framework for sentence-level interpretability of rubric-based scoring that combines model-agnostic Shapley value attributions with rationales generated by large language models (LLMs). Instantiated on the Quality of Feedback dimension of the CLASS framework using the NCTE corpus, the framework enables systematic comparison of fine-tuned pretrained language models (PLMs) and prompted LLMs on both scoring performance and explanation faithfulness. Across 6k annotated transcript segments, fine-tuned PLMs outperform LLMs in prediction accuracy but exhibit label compression toward mid-scale scores. Deletion-based tests show that SHAP identifies sentences that reliably drive model predictions, typically producing larger and more coherent prediction shifts than LLM-generated rationales. Cross-model analyses further reveal that SHAP attributions transfer robustly across architectures, whereas LLM rationales exert limited and inconsistent influence. Overall, the findings demonstrate that SHAP provides more faithful and transferable explanations for rubric-based scoring, and that the proposed framework offers a principled basis for evaluating both scoring models and their explanations in high-stakes educational settings and other rubric-based language assessment tasks.
\end{abstract}

\section{Introduction}
Rubric-based scoring models are increasingly used to automatically evaluate open-ended language tasks, from student essays and peer feedback to clinical notes and classroom transcripts. In these settings, models assign scalar scores on multi-level rubrics that inform teaching, evaluation, and policy decisions, yet most systems provide little insight into why a particular score was assigned. This is especially problematic in high-stakes educational contexts, where stakeholders such as teachers must be able to understand, trust, and contest automated judgments—requirements that are now explicitly reflected in emerging regulatory frameworks such as the EU AI Act~\citep{eu_ai_act_2024}. Therefore, a central challenge emerges: \textbf{how can we trust rubric-based scores produced by opaque, black-box models such as large language models (LLMs) when their internal decision-making is inaccessible and their explanations may be unfaithful?} Recent work suggests that free-form explanations produced by LLMs can be persuasive without faithfully reflecting the underlying computation~\citep{turpin2023languagemodels, ye2022theunreliability}, thus raising a critical question for explainable NLP: \textbf{which parts of a text truly drive a model’s rubric-based score, and how can we evaluate whether an explanation has captured them?}

High-quality feedback is central to teachers’ professional growth, yet providing consistent and individualized feedback is resource-intensive and prone to inconsistency. Recent work demonstrates that automated feedback tools can enhance teachers’ uptake of student ideas by as much as 24\%~\citep{demsky2024canautomated}, showing the promise of NLP-based approaches for supporting teacher development. Classroom teaching quality is thus a prototypical example of a high-stakes, rubric-based judgment where opaque scores are insufficient, and explanations are critical.

Automated scoring of teaching quality dimensions has likewise proven feasible~\citep{hou2024automatedassessment, FUTTERER2026102264}, but scoring alone does not provide insight into the reasoning behind a model’s evaluation. Moving from \textit{what} (the score) to \textit{why} (the reasoning) is essential for generating actionable feedback and fostering user trust. However, whereas LLMs can generate rich, sentence-level rationales, a growing body of work shows that such explanations often fail to reflect the model’s actual decision process~\cite{turpin2023languagemodels, ye2022theunreliability}. Despite the rapid progress of LLMs and transformer-based scoring systems, their decision-making processes often remain opaque.

To bridge this gap, we investigate explainable NLP methods that can reveal which parts of classroom dialogue most strongly influence automated teaching quality assessments. Specifically, we propose a unified framework for sentence-level interpretability of rubric-based teaching quality scores, comparing model-agnostic feature attribution using SHAP~\citep{lundberg2017aunifiedapproach} with LLM-based reasoning to identify aspects of teacher–student interaction that contribute to high- or low-quality feedback. Our study focuses on the \textit{Quality of Feedback} dimension, evaluating the quality of the feedback given by teachers to their students in the classroom, within the \textit{Instructional Support} domain of the Classroom Assessment Scoring System (CLASS) framework~\citep{Pianta2008ClassroomAS}, as providing feedback is a core teaching practice and exhibits a balanced label distribution in our dataset.

We evaluate fine-tuned transformer-based models and LLMs on the NCTE dataset~\citep{demszky-hill-2023-ncte}, containing elementary mathematics classroom transcripts annotated by expert observers. Beyond comparing model performance, our work examines the faithfulness and consistency of different explanation methods by systematically removing sentences highlighted as important, either by SHAP or by LLM-generated rationales, and measuring how these removals change predictions. We further introduce a cross-model evaluation protocol where explanations generated for one model family are used to perturb inputs for the other, allowing us to study whether explanations transfer across architectures or remain model-specific.

Our work contributes to the growing body of research on explainable NLP and reliable LLM rationales in education and other rubric-based assessment settings by:
\begin{enumerate}
    \item Proposing a general framework for sentence-level interpretability of rubric-based scoring models that combines model-agnostic Shapley value attributions with LLM-generated rationales, instantiated on automated teaching quality assessment.
    \item Comparing specialized fine-tuned models and LLM prompting for teaching quality scoring.
    \item Evaluating the faithfulness of SHAP and LLM-based explanations through deletion-based tests, assessing how influential the identified sentences truly are for model predictions.
    \item Introducing cross-model consistency analyses, where LLM-selected sentences are removed and evaluated with fine-tuned models (and vice versa), to probe the alignment of explanation methods across architectures.
    \item Discussing the implications of our findings for the design of transparent, actionable teacher feedback tools and, more broadly, for the use of LLM rationales as explanations in rubric-based educational assessments.
\end{enumerate}

We address the following research questions (RQs):
\begin{itemize}
    \item \textbf{RQ1:} How do fine-tuned transformer-based pretrained language models (PLMs) compare to prompted LLMs in predicting rubric-based teaching quality scores on the Quality of Feedback dimension?
    \item \textbf{RQ2:} How faithful and reliable are SHAP- and LLM-based sentence-level explanations in identifying influential parts of a classroom transcript, as measured by deletion-based changes in predictions?
    \item \textbf{RQ3:} To what extent do explanations transfer across model types, i.e., does removing sentences identified by one model meaningfully affect the predictions of another?
\end{itemize}

\section{Related Work}

\subsection{Explainability Methods for NLP Models}

Explainable NLP methods aim to identify input components that most strongly influence model predictions, a crucial requirement in educational contexts where transparency is essential. Model-agnostic approaches, e.g., LIME~\citep{ribeiro2016whyitrustyou} and SHAP~\citep{lundberg2017aunifiedapproach}, provide local feature attributions by approximating complex classifiers or computing Shapley values. Beyond NLP, SHAP has also been applied in other domains to improve model interpretability, including work that combines SHAP explanations with LLM-generated descriptions to enhance human-understandable rationales~\citep{khediri2024enhancingmachine}. In our work, we use SHAP at a sentence embedding level within a hierarchical PLM architecture, treating sentences as features to obtain document-level attributions that are both computationally tractable and directly actionable for feedback.

For neural text models, attention-based interpretations have been debated due to concerns about whether attention weights reflect causal importance~\citep{jain-wallace-2019-attention, wiegreffe-pinter-2019-attention}. To address these limitations, research increasingly distinguishes plausibility from faithfulness~\citep{jacovi-goldberg-2020-towards}. Deletion- and perturbation-based evaluation, i.e., removing influential input elements and observing prediction changes, provides a more direct measure of explanation faithfulness~\citep{deyoung-etal-2020-eraser}. We adopt this perspective and extend it to a cross-model setting: explanations are evaluated not only with respect to their source model, but also by measuring how they perturb predictions of alternative architectures. Furthermore, our work contributes by applying sentence-level SHAP in a hierarchical PLM setting and evaluating its faithfulness through systematic deletion tests.

Recent work in educational assessment has also explored the use of SHAP to interpret rubric-based scoring models. For example, \citet{boulanger2020shaped} apply SHAP to automated essay scoring to quantify the contribution of linguistic features to rubric-level predictions, enabling both local and global interpretability of model behavior. Similarly, \citet{kumar2020explainable} demonstrate that combining deep learning with SHAP can expose the decision-making process of rubric-based scoring systems and support the generation of fine-grained, formative feedback aligned with pedagogical criteria. These findings highlight the potential of SHAP not only as a diagnostic tool for model transparency, but also as a bridge between model predictions and human-interpretable rubric constructs in educational settings.

\subsection{Faithfulness and Reliability of Model Explanations in LLMs}

LLMs are increasingly producing free-form rationales and structured justifications for their predictions. However, a growing body of work suggests that these explanations may not accurately reflect the underlying computation. Chain-of-thought rationales can be unfaithful even while producing correct answers~\citep{turpin2023languagemodels}, and explanations in few-shot prompts frequently exhibit inconsistencies or hallucinations~\citep{ye2022theunreliability}. Structured prompting can improve reliability, but challenges remain~\citep{ayala-bechard-2024-reducing}.

Recent approaches have proposed adapting Shapley-based methods to LLMs~\citep{mohammadi_explaining_2024}, although computational constraints limit their practical use. Despite the increased adoption of LLMs in educational settings, the faithfulness of their rationales has not been systematically evaluated relative to established attribution methods such as SHAP, especially on long, naturalistic transcripts and multi-level rubrics. Our work addresses this gap by comparing LLM-generated sentence rankings against PLM-based SHAP attributions using matched deletion-based faithfulness tests and cross-model robustness analysis, thereby providing empirical evidence on when LLM rationales align with model behavior and when they diverge.

\subsection{Automated Scoring and Teacher Feedback in Educational Settings}

Recent efforts have explored automated approaches to analyze classroom instruction and support teacher learning. The NCTE dataset~\citep{demszky-hill-2023-ncte} has facilitated large-scale research on evaluating teacher uptake of student ideas~\citep{demsky2024canautomated}, and on leveraging models such as ChatGPT for instructional scoring and feedback~\citep{wang-demszky-2023-chatgpt}. More broadly, researchers have begun to validate automated assessments of teaching quality dimensions using multimodal approaches, including audio, video, and text features, combining embeddings with LLM-generated scores~\citep{hou2024automatedassessment, hou2025multimodal, FUTTERER2026102264}. In particular,~\citet{hou-etal-2025-llm} evaluate LLM-based multimodal models for classroom assessment by comparing their predictions against human annotations, providing evidence that such models can approximate human judgments of teaching quality.

In parallel, NLP has long been applied to educational assessment tasks, including essay scoring and discussion analysis. Recent work shows that LLMs and PLMs can approximate human rubric-based judgments across multiple writing dimensions~\citep{sessler2025can}, while neural models have been used to assess the quality of classroom discussions or participation~\citep{tran2023utilizingnatural}. LLMs are increasingly used to provide pedagogically aligned feedback to learners~\citep{MEYER2024100199}. Our work situates itself within this line of research, but shifts the focus from \emph{predictive performance} to \emph{explanation quality}. More specifically, we compare fine-tuned PLMs and instruction-tuned LLMs for scoring a specific CLASS dimension (i.e., Quality of Feedback). More importantly, we introduce a general framework for evaluating sentence-level explanations for rubric-based scores.

\section{Methods}
We cast teaching quality assessment as a general rubric-based text scoring problem with model-agnostic sentence-level explanations and instantiate this framework using both PLMs and instruction-tuned LLMs.

\begin{figure*}
    \centering
    \includegraphics[width=0.91\linewidth]{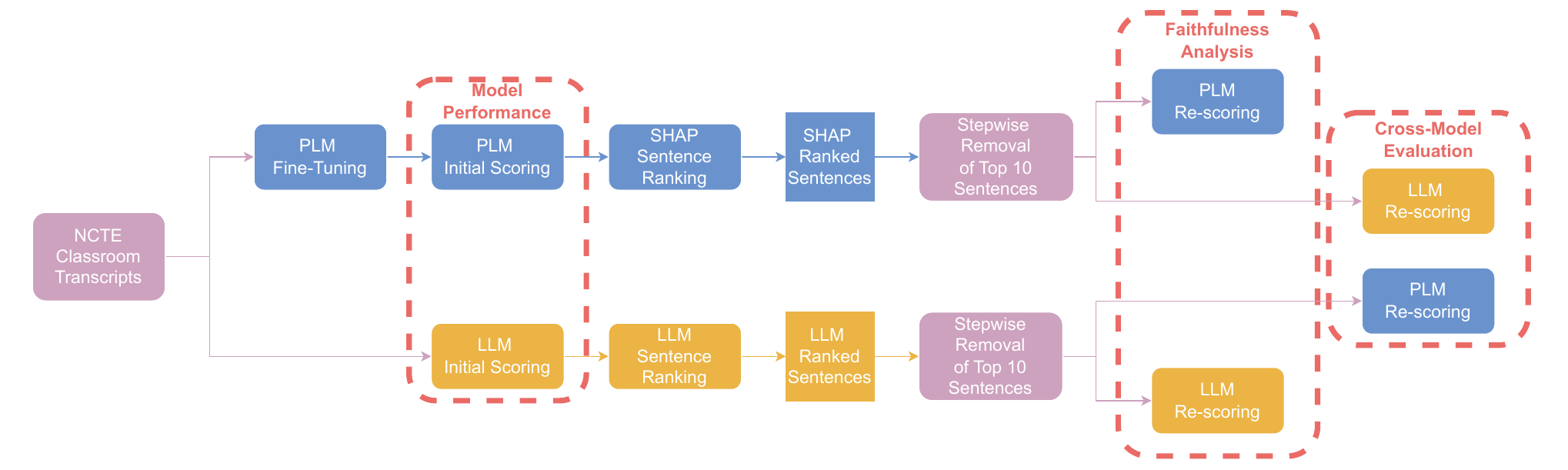}
    \caption{Overview of the proposed framework. The top branch ({\color{f_blue}blue}) shows the PLM pipeline, including fine-tuning, scoring, SHAP-based sentence ranking, and sentence removal with re-scoring. The bottom branch ({\color{f_yellow}yellow}) depicts the corresponding LLM pipeline with prompted scoring and ranking. The three experimental settings are indicated by {\color{f_red}dotted red boxes}.}
    \label{fig:methoddesign}
\end{figure*}
\subsection{Dataset}
To instantiate our interpretability framework in an educational setting, we use the NCTE dataset~\citep{demszky-hill-2023-ncte}, comprising over 1,600 transcripts of 45–60 minute elementary mathematics lessons. Lessons are segmented into 15-minute units, producing 6,005 segments annotated using the CLASS framework~\citep{Pianta2008ClassroomAS}, a tool that measures the quality of interactions between teachers and students to assess teaching. CLASS comprises three domains—\textit{Emotional Support}, \textit{Classroom Management}, and \textit{Instructional Support}—covering thirteen dimensions. We focus on the \textit{Quality of Feedback} (QoF) dimension within the Instructional Support domain.

QoF measures the extent to which teachers provide meaningful feedback, scaffold student thinking, prompt metacognition, elaborate on student responses, and clarify misunderstandings. Each segment receives a QoF rating on a 1–7 scale, where 1–2 indicates low-quality or absent feedback, 3–5 reflects moderate or inconsistent feedback, and 6–7 represents consistently high-quality feedback.

QoF is chosen for three reasons: (1) compared to other CLASS dimensions, its label distribution is less skewed (mean of 4.21, standard deviation of 1.13, with 81\% of ratings in the 3–5 range), making it more suitable for supervised learning; (2) QoF is a core instructional practice strongly linked to student learning gains~\citep{hattie2007thepoweroffeedback}; and (3) many indicators of feedback quality (e.g., probing questions, elaborations, scaffolding) manifest clearly in text transcripts~\citep{demszky-hill-2023-ncte}, making QoF particularly appropriate for sentence-level interpretability analysis.

We adopt an 80/20 data split, resulting in 4,775 segments for training and 1,230 for testing. Transcripts belonging to the same class were kept in the same split, and we stratified the data based on label distribution. The training split is used exclusively for fine-tuning PLMs, whereas the test split is used to evaluate both PLMs and LLMs, as well as to conduct all interpretability experiments. Within the test split, 29.3\% of the sentences are student utterances, 69.9\% are teacher utterances, and 0.8\% were utterances that could not be assigned to a speaker. The dataset provides one expert annotation per teaching quality dimension, precluding the assessment of interrater agreement, which we treat as ground truth for both scoring and evaluation. The dataset does not include human-annotated sentence-level rationales or evidence.

\subsection{Models}
Within our framework, we instantiate the scoring model $f$ using two classes of architectures: PLMs and instruction-tuned LLMs. We compare these two families because PLMs represent the standard supervised approach for rubric-based scoring, requiring task-specific fine-tuning, whereas LLMs offer strong zero- or few-shot capabilities without additional training. This contrast allows us to examine differences in scoring performance but also how explanation methods behave across models with fundamentally different training paradigms, capacities, and levels of transparency.

\paragraph{Pretrained Language Models.}
We fine-tuned \textit{BERT}~\citep{devlin-etal-2019-bert}, \textit{ALBERT}~\citep{lan2020albertlitebertselfsupervised}, \textit{RoBERTa}~\citep{liu2019robertarobustlyoptimizedbert}, and \textit{DeBERTa~V3}~\citep{he2023debertav3improvingdebertausing}, using both base and large variants. PLMs operate on transcript segments at the sentence level: each sentence was encoded using the model’s \texttt{[CLS]} representation, and a trainable attention layer computed an attention-weighted document embedding. A linear regression head predicted a single scalar QoF score. We used a maximum sentence length of 128 tokens and a maximum of 263 sentences per document, which encompasses 98\% of the sentence lengths and 90\% of the document lengths without truncation. During fine-tuning, models were optimized using mean squared error; additional hyperparameters are listed in App.~\ref{app:hyperparams}. After fine-tuning, we applied SHAP to the document-level regression output, treating each sentence embedding as a separate feature. SHAP returned one Shapley value per sentence, representing its estimated contribution to the predicted score.

\paragraph{Large Language Models.}
For LLMs, we used instruction-tuned variants of Llama~3.1 (8B, 70B)~\citep{grattafiori2024llama3herdmodels}, Mixtral (8$\times$7B, 8$\times$22B)~\citep{jiang2024mixtralexperts}, Qwen~3 (4B, 30B, 235B)~\citep{yang2025qwen3technicalreport}, and Mistral (Small, Small~24B)~\citep{jiang2023mistral7b}. Only open-source LLMs were selected, as they can be deployed locally and therefore mitigate the privacy concerns associated with datasets such as classroom transcripts. LLMs performed two tasks: (1) QoF scoring using a few-shot prompt introducing the task and showing examples, and (2) sentence ranking using a zero-shot prompt. For ranking, LLMs were used as sentence-level evidence selectors, rather than to free-form explanations. We provided transcripts segmented and numbered by sentence and requested a list of ten sentence indices corresponding to the most influential sentences. This ensured alignment with the PLM sentence boundaries and prevented models from altering sentence content. Some outputs contained invalid indices or fewer than ten items; in such cases, the system retried up to ten times. Prompts are shown in App.~\ref{app:prompts}. All models were evaluated with deterministic decoding, and all local inference used 4-bit \texttt{nf4} quantization via BitsAndBytes.

\subsection{Interpretability Methods}
We view rubric-based teaching quality assessment as a complex text scoring problem. More specifically, given an input transcript $x$ and a model $f$, the model outputs a scalar score $f(x)$ on a fixed rubric. In this work, we use the term `explanations' to refer to sentence-level evidence, i.e., subsets of transcript sentences identified as most influential for a model’s prediction, following common usage in extractive explanation methods such as SHAP. An explanation method $E$ maps $x$ and $f$ to a ranked list of textual units (here, sentences) that are claimed to be most influential for the score. To evaluate the faithfulness of such explanations, we adopt a deletion-based protocol: we progressively remove the top-$k$ units selected by $E$ (with $k = 10$ in all experiments), recompute $f(x)$, and measure the change in predictions. Larger changes indicate that the explanation has successfully identified text that the model relies on. To study cross-model consistency, we extend this protocol to pairs of models $f$ and $g$, apply explanations obtained from one model to perturb the other and compare the resulting prediction shifts. In this paper, we instantiate $f$ as fine-tuned PLMs and $g$ as instruction-tuned LLMs, with explanations $E$ provided either by sentence-level SHAP attributions or by LLM-generated sentence rankings.

\subsection{Experiments}
To evaluate our proposed sentence-level interpretability evaluation framework, we conduct three sets of experiments. The framework combines sentence-level explanation generation (via SHAP or LLM-based ranking) with faithfulness testing through sentence deletion and robustness analysis via cross-model transfer. Each experiment targets one aspect of this framework across model families.
\textbf{(1) Model performance for scoring (RQ1):} We evaluate PLMs and LLMs on the transcript segment scoring task.
\textbf{(2) Faithfulness analysis (RQ2):} We compute score differences ($\Delta$) after each single-sentence deletion to assess whether a model’s own explanations meaningfully affect its predictions, as described by the equation:
\begin{equation}
    \Delta_i = f(x_{-r_{i-1}}) - f(x_{-r_i}) 
    \label{eq:delta_i}
\end{equation}

where $r_i$ denotes the i-th ranked sentence and $x_{-r_i}$ the input after its removal. All models employ the same sentence-splitting procedure, and deletions are made in accordance with the ranking order. When fewer than ten sentences exist or when deletions result in an empty transcript, models are prompted with an empty input. This only happened for 17 (1.3\%) out of 1,230 transcript segments. In addition to deletion-based evaluation, we also quantify the alignment between different explanation methods. For each transcript, we compute the Jaccard similarity between the sets of top-$k$ sentence indices (here $k=10$) selected by SHAP and by each LLM, i.e., the overlap between the top-k sentence IDs selected by each method for a given transcript, and the Spearman rank correlation, which is computed only when both methods produce valid rankings over the same set of sentences, excluding cases with missing or invalid indices where a consistent ranking is not defined. These metrics capture how often explanation methods highlight the same textual evidence, independently of their causal impact on predictions.
\textbf{(3) Cross-model evaluation (RQ3):} We examine whether sentence-level explanations generalize across models by applying sentence deletion based on SHAP and LLM rankings to the opposite model family. We select six representative models for this analysis: BERT~large, DeBERTa~V3~large, and ALBERT~base (PLMs) and Qwen~3~235B, Mistral Small, and Llama~3~8B (LLMs). We select these representative models based on their sensitivity to sentence removal, measured by the cumulative prediction change ($\Delta$) over the top-k deletions. Specifically, we choose models with the highest, lowest, and intermediate $\Delta$ values to capture a range of faithfulness behaviors.

We report mean absolute error (MAE) and mean squared error (MSE). For constant baselines, we use the median prediction for MAE and the mean prediction for MSE, corresponding to the respective risk minimizers. All experiments use identical sentence segmentation, deletion rules, and deterministic LLM decoding.

\section{Results And Discussion}
\label{sec:resultsanddiscussion}

\subsection{Model Performance for Scoring}

Table~\ref{tab:originalscores} reports Mean Absolute Error (MAE) and Mean Squared Error (MSE) for all models. For PLMs, results are shown before and after fine-tuning, whereas for LLMs, a single score is reported based on the few-shot scoring prompt (see App.~\ref{app:prompts}).

\begin{table}[!ht]
    \centering
    \scalebox{0.725}{
    \begin{tabular}{l c c c c}
        \hline
        \textbf{Model} & \textbf{MAE} & \textbf{MSE} & \textbf{MAE} & \textbf{MSE} \\
        \hline
        Constant Baseline & 0.96 & 1.35 & -- & -- \\
        \multicolumn{5}{l}{} \\
    
        \hline
        & \multicolumn{2}{c}{\textbf{Non-Fine-Tuned}} & \multicolumn{2}{c}{\textbf{Fine-Tuned}} \\
        \cmidrule(lr){2-3} \cmidrule(lr){4-5}
        ALBERT base        & 4.34 & 20.12 & 0.98 & 1.34 \\
        ALBERT large       & 4.22 & 19.10 & 0.99 & 1.45 \\
        BERT base          & 4.58 & 22.29 & 0.98 & 1.36 \\
        BERT large         & 4.45 & 21.07 & 0.97 & 1.34 \\
        DeBERTaV3 base     & 4.08 & 17.97 & 1.00 & 1.56 \\
        DeBERTaV3 large    & 3.66 & 14.65 & \textbf{0.96} & \textbf{1.31} \\
        RoBERTa base       & 4.33 & 20.08 & 0.97 & 1.34 \\
        RoBERTa large      & 3.27 & 12.01 & 0.97 & 1.37 \\
        \multicolumn{5}{l}{} \\
        \hline
        Llama 3.1 8B Instruct       & 1.63 & 3.98 & -- & -- \\
        Llama 3.1 70B Instruct      & 1.98 & 5.32 & -- & -- \\
        Mistral Small Instruct      & \textbf{1.02} & \textbf{1.78} & -- & -- \\
        Mistral Small 24B Instruct  & 1.75 & 4.28 & -- & -- \\
        Mixtral 8x7B Instruct       & 1.39 & 2.85 & -- & -- \\
        Mixtral 8x22B Instruct      & 1.21 & 2.41 & -- & -- \\
        Qwen3 4B Instruct           & 1.18 & 2.29 & -- & -- \\
        Qwen3 30B A3B Instruct      & 1.56 & 3.59 & -- & -- \\
        Qwen3 235B A22B Instruct    & 1.67 & 4.16 & -- & -- \\
        \hline
    \end{tabular}

    }
    \caption{Mean Absolute Error (MAE) and Mean Squared Error (MSE) for PLMs and LLMs. PLM results are reported separately for non-fine-tuned and fine-tuned models, while LLM results correspond to prompted inference without task-specific training.}
    \label{tab:originalscores}
\end{table}

For PLMs, fine-tuning yields a substantial and expected performance improvement. While non-fine-tuned models show MAEs above 4.0, all fine-tuned PLMs achieve uniformly low errors (MAE 0.96–1.00; MSE 1.31–1.56), comparable to the constant baseline. Performance differences among fine-tuned PLMs are minimal ($\Delta=0.04$ MAE, $\Delta=0.25$ MSE), indicating comparable behavior across architectures. The best-performing model, DeBERTaV3~large, achieves an MAE of 0.96 and an MSE of 1.31. Given the 1–7 scoring scale, an MAE of \textasciitilde 1 corresponds to an average error of approximately one rubric point, indicating that predictions are typically within one level of the expert annotation. The corresponding MSE values (\textasciitilde 1.3–1.5) reflect relatively small squared deviations, consistent with the low MAE.

Despite their strong numerical performance, fine-tuned PLMs exhibit limited label coverage. For DeBERTaV3~large, the mean predicted score is 4.14 ($\sigma=0.16$), closely matching the dataset average of 4.22, but predictions never fall below 2.03 or exceed 5.89, meaning that extreme labels (1 and 7) are never predicted (see App.~\ref{app:pred_stats}). This behavior is consistent across all fine-tuned PLMs, with some models (e.g., ALBERT and RoBERTa variants) effectively collapsing to a narrow mid-range of labels (3–5). This pattern is likely driven by strong label imbalance at the extremes of the QoF scale, indicating high sensitivity to the underlying data distribution.

LLMs show weaker overall scoring performance than fine-tuned PLMs. The best-performing LLM, Mistral Small Instruct, achieves an MAE of 1.02 and an MSE of 1.78, which remains higher than both the best PLM and the constant baseline. In contrast to PLMs, performance variability across LLMs is substantially larger ($\Delta=0.96$ MAE, $\Delta=3.54$ MSE), reflecting notable differences across models. However, most LLMs produce predictions spanning the full 1–7 score range. For Mistral Small Instruct, the mean predicted score is 4.37 ($\sigma=0.77$), indicating a much broader dispersion than observed for PLMs.

Overall, these results highlight a trade-off between accuracy and flexibility. Fine-tuned PLMs achieve substantially higher scoring accuracy but are very sensitive to training data distribution, whereas LLMs provide wider score distributions at the cost of reduced accuracy. In educational settings, LLMs offer practical advantages due to their out-of-the-box usability and lack of task-specific training requirements, but incur higher computational costs and lower predictive reliability. In response to RQ1, fine-tuned PLMs outperform prompted LLMs in accuracy, while LLMs better preserve score variability.

\subsection{Faithfulness Analysis}

The second experiment evaluates the faithfulness of SHAP- and LLM-based sentence importance rankings. For SHAP, the number of selected sentences is deterministically fixed as the minimum of ten and the number of sentences in each transcript segment. In contrast, LLMs show limited controllability: even when prompted to return exactly ten sentences, they produce more or fewer often. Models such as Mixtral~8$\times$7B and Qwen~3~235B deviate most strongly from the expected average of 9.931 sentences, typically returning fewer sentences despite up to ten retry attempts for malformed outputs. Llama~3.1~8B is the only model that, on average, returns more sentences than requested and comes closest to the target value (see App.~\ref{app:number_of_sents}).

\begin{table}
    \centering
    \scalebox{0.8}{
    \begin{tabular}{llc}
        \toprule
        \textbf{Group} & \textbf{Model} & \boldmath{$\mathbf{\overline{\Delta}}$} \\
        \midrule
        \multirow{9}{*}{\textbf{PLMs}}
         & ALBERT base        & 0.0219 \\
         & ALBERT large       & 0.0172 \\
         & BERT base          & 0.0256 \\
         & BERT large         & \textbf{0.0329} \\
         & DeBERTaV3 base     & 0.0242 \\
         & DeBERTaV3 large    & 0.0049 \\
         & RoBERTa base       & 0.0053 \\
         & RoBERTa large      & 0.0119 \\
        \midrule
        \multirow{9}{*}{\textbf{LLMs}}
         & Llama 3.1 8B Instruct        & 0.0174 \\
         & Llama 3.1 70B Instruct       & 0.0090 \\
         & Mistral Small Instruct       & 0.0033 \\
         & Mistral Small 24B Instruct   & 0.0123 \\
         & Mixtral 8x7B Instruct        & 0.0121 \\
         & Mixtral 8x22B Instruct       & 0.0199 \\
         & Qwen3 4B Instruct            & 0.0211 \\
         & Qwen3 30B A3B Instruct       & 0.0174 \\
         & Qwen3 235B A22B Instruct     & \textbf{0.0388} \\
        \bottomrule
    \end{tabular}
    }
    \caption{Average consecutive performance change $\overline{\Delta}$ across sentence removals for PLMs and LLMs.}
    \label{tab:avg_delta}
\end{table}

These results highlight the inherent variability and limited controllability of LLM outputs, even under structured prompting and multiple retries. This unpredictability represents a practical limitation when deploying LLMs in educational systems, where reliability and strict adherence to output formats are critical. Systems that incorporate LLMs must therefore be explicitly designed to handle such inconsistencies if these models are to be used effectively in real-world educational settings.

We then remove one sentence at a time from the top ten ranked sentences and re-score each transcript segment using the same model that produced the ranking. Table~\ref{tab:avg_delta} reports the average prediction change after each consecutive removal, denoted as $\overline{\Delta}$, and calculated as follows:
\begin{equation}
    \overline{\Delta} = \frac{1}{k} \sum_{i=1}^{k} \Delta_i
\end{equation}
where $k=10$ is the number of sentences deleted, and $\Delta_i$ is calculated as described in Eq.~\ref{eq:delta_i}. Among PLMs, BERT models are most sensitive to sentence removal, with BERT~large exhibiting the highest average change ($\overline{\Delta}=0.0329$). For LLMs, the largest change is observed for Qwen~3~235B ($\overline{\Delta}=0.0388$), while all other LLMs show $\overline{\Delta}$ values below 0.02. Overall, this indicates that both model families can identify influential sentences, though PLMs do so more consistently, while LLM behavior is more variable (see Section~\ref{sec:crossmodeleval}).

Beyond faithfulness, we observe largely similar patterns in the types of sentences selected by PLMs and LLMs. Both predominantly prioritize teacher-authored utterances, consistent with the QoF rating dimension: LLMs select teacher utterances in 79.5\% of cases and student utterances in 19.5\%, while PLMs show a comparable distribution (74.0\% teacher, 24.7\% student). Despite this similarity, alignment between SHAP- and LLM-based explanations remains consistently low. Across nine LLMs, the mean Jaccard similarity between the indices of the top-10 sentence sets is 0.085, corresponding to an overlap of roughly one to two sentences per transcript, while the mean Spearman rank correlation is 0.062, indicating weak agreement in sentence importance ordering. These trends are consistent across model families and sizes, suggesting that explanation alignment is driven primarily by the explanation method rather than model scale.

\begin{figure}
    \centering
    \includegraphics[width=\linewidth]{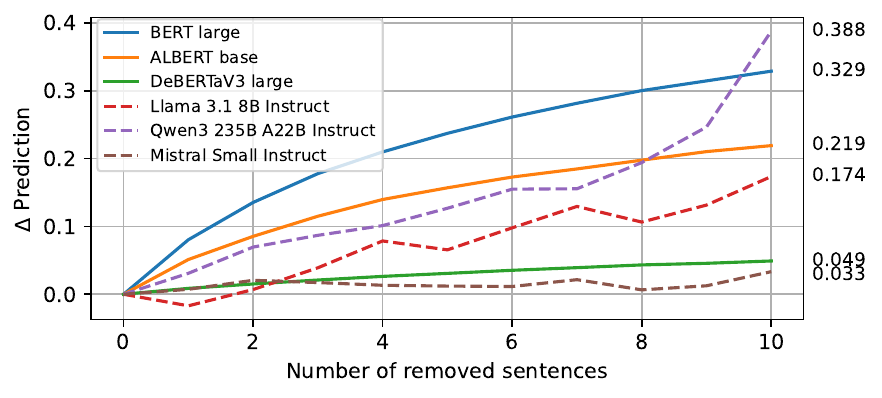}
    \caption{Prediction change $\Delta$ under progressive sentence removal for selected PLMs and LLMs. Sentences are chosen using SHAP (solid lines) or LLM-based rankings (dashed lines), and inputs are re-scored by the same model; final changes are shown on the right.}
    \label{fig:delta_selected_three}
\end{figure}

Fig.~\ref{fig:delta_selected_three} illustrates prediction changes under progressive sentence removal for selected PLMs and LLMs. For completeness, App.~\ref{app:full_sent_removal} reports results for all models, App.~\ref{app:sent_examples} provides representative sentence examples, and App.~\ref{sec:alignment_appendix} contains the full alignment statistics. Negative $\Delta$ values indicate that removing a sentence increases the predicted score, implying that the sentence contributed negatively to the model’s assessment. This behavior is consistent with the rubric, where low-quality feedback instances are expected to reduce the overall score.

Together, these results answer RQ2: SHAP-based sentence rankings are more faithful for PLM scorers, whereas LLM-generated rationales induce smaller and often unstable prediction changes, even for the models that produce them.

\subsection{Cross-Model Evaluation}
\label{sec:crossmodeleval}

For cross-model evaluation, we select three PLMs and three LLMs representing high, medium, and low faithfulness in the single-model deletion analysis. Specifically, we use \textbf{BERT~large} (most affected), \textbf{ALBERT~base} (moderately affected), and \textbf{DeBERTaV3~large} (least affected) among PLMs, and \textbf{Qwen3~235B}, \textbf{Llama~3.1~8B}, and \textbf{Mistral~Small} as corresponding LLMs. Their sentence-removal trajectories under self-generated explanations are shown in Fig.~\ref{fig:delta_selected_three}, with full results in App.~\ref{app:full_sent_removal}.

\begin{figure*}
    \centering
    \includegraphics[width=0.91\linewidth]{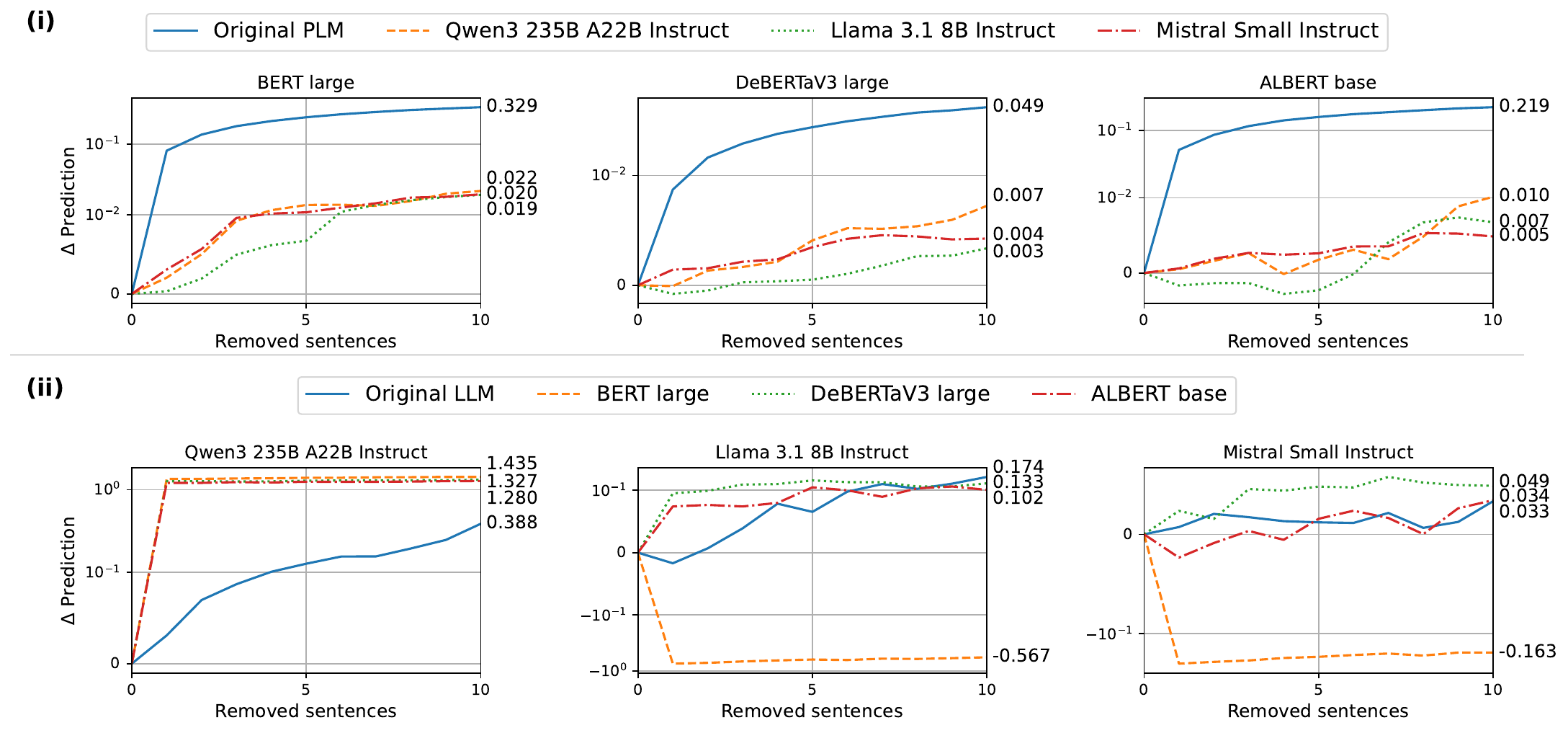}
    \caption{Prediction change $\Delta$ under progressive sentence removal for selected PLMs and LLMs. Panel (i) shows PLM re-scoring after removing sentences selected by SHAP (solid) or LLM-based rankings (dashed/dotted); panel (ii) shows LLM re-scoring after removing sentences selected by the LLM itself (solid) or by SHAP from fine-tuned PLMs (dashed/dotted).}
    \label{fig:delta}
\end{figure*}

To assess transfer from LLMs to PLMs, we apply LLM-generated sentence rankings to PLMs and re-score the perturbed inputs. Fig.~\ref{fig:delta}~(i) compares these results with the baseline condition where PLMs are perturbed using SHAP-selected sentences. Across all models, removing LLM-ranked sentences produces substantially smaller prediction shifts than removing SHAP-ranked sentences, even for PLMs that are weakly sensitive to SHAP. Moreover, deletion trajectories induced by LLM rationales are often non-monotonic, with predictions fluctuating as additional sentences are removed, indicating limited alignment with the features PLMs rely on. Among LLMs, \textbf{Qwen3~235B} shows the strongest cross-model transfer, yielding the largest and most stable perturbations, though still markedly weaker than those induced by SHAP.

We then consider the reverse direction, applying PLM-derived SHAP explanations to LLM scoring. As shown in Fig.~\ref{fig:delta}~(ii), removing the single most influential SHAP-ranked sentence frequently causes a large immediate shift in LLM predictions, followed by stabilization in subsequent deletions. This effect is most pronounced for PLM–LLM pairs that are highly sensitive to sentence removal. The consistent “first-step jump” suggests that SHAP identifies sentence-level features that are relevant not only to PLMs but also to LLMs, despite architectural and training differences.

Overall, these results indicate that PLMs and LLMs rely on different sentence-level evidence. While LLM rationales often surface intuitively relevant content, they do not reliably capture the features driving PLM predictions, regardless of model sensitivity. In contrast, SHAP explanations consistently induce larger, more stable, and more coherent prediction shifts both within PLMs and when transferred to LLMs, demonstrating superior faithfulness for sentence-level interpretability.

This difference is likely influenced by architectural factors: PLMs are trained with sentence-level representations, whereas LLMs operate primarily at the token level. Nevertheless, sentence-ranking attributions remain a useful tool for improving the interpretability of rubric-based scoring, particularly when LLMs are used as classifiers. Addressing RQ3, the cross-model evaluation shows that SHAP explanations generalize across architectures, while LLM rationales transfer poorly and appear unreliable as general-purpose explanations.

\subsection{Ablation Study}
To control for potential structural artifacts, we compare against a random sentence-deletion baseline matched for sentence length.  Table~\ref{tab:avg_delta_random} shows a comparison of the average prediction change between removing ranked sentences and removing random sentences. This baseline yields near-zero prediction changes between consecutive removals, indicating that the larger effects observed for SHAP- and LLM-based rankings are not explained by sentence length or generic perturbation, but by the identification of influential content.

\begin{table}
    \centering
    \scalebox{0.773}{
    \begin{tabular}{llcc}
        \toprule
        \textbf{Group} & \textbf{Model} &  \textbf{Ranked} & \textbf{Random} \\
        & & \boldmath{$\mathbf{\overline{\Delta}}$} & \boldmath{$\mathbf{\overline{\Delta}}$} \\
        \midrule
        \multirow{3}{*}{\textbf{PLMs}}
         & ALBERT base        & 0.0219 & 0.0075 \\
         & BERT large         & 0.0329 & 0.0082 \\
         & DeBERTaV3 large    & 0.0049 & -0.0064 \\
        \midrule
        \multirow{3}{*}{\textbf{LLMs}}
         & Llama 3.1 8B Instruct      & 0.0174 & -0.0076 \\
         & Mistral Small Instruct     & 0.0033 & 0.0016 \\
         & Qwen3 235B A22B Instruct   & 0.0388 & -0.0036 \\
        \bottomrule
    \end{tabular}
    }
    \caption{Average prediction change ($\overline{\Delta}$) comparing ranked and random sentence removal for PLMs and LLMs, serving as a baseline for explanation faithfulness.}
    \label{tab:avg_delta_random}
\end{table}

\section{Conclusion}
We propose a general framework for evaluating the faithfulness of sentence-level explanations in rubric-based scoring, systematically contrasting Shapley value attributions with LLM-generated rationales based on their causal impact on model predictions. Applied to the QoF dimension, the framework shows that fine-tuned PLMs outperform prompted LLMs in scoring accuracy (RQ1), although PLMs exhibit label compression, whereas LLMs provide broader but less precise predictions. Despite the relatively low error values, there remains room for improvement in prediction accuracy, particularly for extreme score ranges that are underrepresented in the data. Deletion-based tests demonstrate that SHAP explanations are substantially more faithful than LLM rationales (RQ2), and cross-model evaluations reveal that SHAP-selected sentences transfer more robustly across architectures, whereas LLM rationales exert limited influence on PLM predictions (RQ3). Overall, the results suggest that current LLM rationales are unreliable as faithful justifications for rubric-based scores, whereas Shapley-based attributions provide a more stable foundation for transparent and actionable automated assessment. More broadly, our findings suggest that the proposed framework offers a robust and extensible way to evaluate scoring models and their explanations in high-stakes settings such as education and can be readily applied to other rubric-based language assessment tasks (e.g., essay scoring, peer feedback quality, or clinical note evaluation).

\section*{Limitations}
A primary limitation of this work is the dataset's size and label distribution. Although we selected the least skewed CLASS dimension, only 19\% of the labels fall outside the 3–5 range across 6k transcript segments. This imbalance likely contributes to the observed label compression in fine-tuned PLMs, limiting model performance at the extremes of the scale. Future work could explore data augmentation or synthetic data generation to improve coverage of underrepresented classes. In addition, our experiments focus solely on the Quality of Feedback dimension, and it remains unclear how well the findings generalize to other CLASS dimensions, particularly those that are less discourse-driven, such as Productivity within the Classroom Management domain.

Our analysis is further restricted to text-only transcripts. CLASS scoring in practice relies on rich, multimodal cues, including prosody, timing, and visual interactional signals, which are currently ignored. Moreover, each transcript segment is annotated by a single expert, preventing assessment of inter-rater reliability and leaving open the possibility of annotation noise and subjective bias. Whereas Quality of Feedback is primarily teacher-centered, a substantial portion of the transcripts consists of student utterances, which were also frequently selected by the sentence-ranking methods. Future work should investigate how explicitly filtering or modeling student contributions affects both scoring and the interpretability of results.

Finally, the deletion-based faithfulness protocol perturbs the natural discourse structure of classroom interaction and may alter pragmatic meaning and speaker intent, limiting its ability to reflect true causal influence. This disruption may disproportionately affect LLM-based scoring and explanations, as LLMs are trained on coherent sentence generation and strongly rely on intact discourse structure and pragmatic flow. Additionally, we do not directly evaluate whether the sentences identified by the ranking methods align with the underlying rubric constructs (i.e., whether they are construct-relevant). Incorporating human judgments on a subset of transcripts and sentence rankings would provide a more direct assessment of this alignment and is an important direction for future work. Finally, we also observed notable reliability issues in LLM sentence-ranking outputs, which required strict prompt engineering and external sentence segmentation to enforce structured predictions, thereby constraining the natural expressive capacity of LLM-based explanations. Nevertheless, we observe a clear asymmetry: removing SHAP-selected sentences leads to substantially larger changes in predicted scores than removing LLM-selected sentences, or randomly selected sentences. This suggests that the observed differences in faithfulness cannot be attributed solely to discourse disruption, but rather reflect differences in how well each method identifies sentences that are truly influential for the model’s predictions.

\section*{Ethical considerations}

A central ethical concern in automated rubric-based scoring is the risk of bias amplification from training labels. Models trained on human annotations may inherit subjective judgments or structural biases and propagate them at scale~\citep{barocas2016big,mehrabi2021survey}. In educational contexts, such effects are particularly concerning, as algorithmic assessment systems can reproduce or exacerbate existing inequalities~\citep{nguyen2023ethical}.

Accordingly, automated scoring systems should not replace human raters in high-stakes settings, where unchecked deployment may create self-reinforcing feedback loops of biased predictions. Instead, these systems should be positioned as self-assessment or decision-support tools that augment, rather than replace, professional judgment, consistent with established principles for ethical and human-centered AI in education~\citep{holmes2022ethics,nguyen2023ethical}.

Relatedly, there is a risk that model predictions and explanations may be interpreted as objective ground truth. In educational settings, this is particularly problematic, as model-generated feedback can influence teaching practices, evaluations, and institutional decision-making. This underscores the critical need for transparency: educators and stakeholders must be able to understand how and why a model arrives at a given score to appropriately contextualize and challenge its outputs. To mitigate misuse, it is essential to communicate that both scores and explanations are probabilistic and imperfect, and that interpretability methods are intended to support teacher reflection and informed decision-making rather than to serve as authoritative or definitive assessments.

Privacy and data protection are also key ethical considerations. Although the NCTE transcripts used in this work are anonymized, classroom interactions inherently involve sensitive information about teachers and minors. Any real-world deployment of similar systems must ensure strict safeguards for data security, informed consent, and compliance with regulations governing the processing of educational data.

Furthermore, the use of LLM-generated rationales introduces additional risks. LLMs are known to hallucinate, produce inconsistent outputs, and lack transparent decision processes, which makes their explanations particularly problematic in high-stakes settings such as education. Limited reproducibility further complicates auditing and accountability. These factors reinforce the need for caution when using LLM rationales as justifications for automated judgments, and motivate the emphasis of this work on faithfulness-based evaluation. In particular, explanation methods should be evaluated not only for plausibility but also for their causal impact on model predictions, and systems should avoid presenting unvalidated LLM rationales as authoritative interpretations of teaching practice.

\paragraph{Use of AI Assistance.}

We used AI assistance tools (ChatGPT, and GitHub Copilot) to aid in rewriting code, and paraphrasing. All AI-generated content was thoroughly reviewed and verified by the authors. AI was not used to generate new research ideas or original findings; rather, it served as a support tool to improve clarity, efficiency, and organization. In accordance with ACL guidelines, our use of AI aligns with permitted assistance categories, and we have transparently reported all relevant usage in this paper. While AI contributed to enhancing the quality of the work, no direct research outputs are the result of AI assistance.

\bibliography{custom}

@article{demsky2024canautomated,
author = {Dorottya Demszky and Jing Liu and Heather C. Hill and Dan Jurafsky and Chris Piech},
title ={Can Automated Feedback Improve Teachers’ Uptake of Student Ideas? Evidence From a Randomized Controlled Trial in a Large-Scale Online Course},
journal = {Educational Evaluation and Policy Analysis},
volume = {46},
number = {3},
pages = {483-505},
year = {2024},
doi = {10.3102/01623737231169270},
URL = {  
        https://doi.org/10.3102/01623737231169270
},
eprint = { 
        https://doi.org/10.3102/01623737231169270
}
}

@article{hou2025multimodal,
  title={Multimodal Assessment of Classroom Discourse Quality: A Text-Centered Attention-Based Multi-Task Learning Approach},
  author={Hou, Ruikun and B{\"u}hler, Babette and F{\"u}tterer, Tim and Bozkir, Efe and Gerjets, Peter and Trautwein, Ulrich and Kasneci, Enkelejda},
  journal={arXiv preprint arXiv:2505.07902},
  year={2025}
}

@InProceedings{hou2024automatedassessment,
    author="Hou, Ruikun
    and F{\"u}tterer, Tim
    and B{\"u}hler, Babette
    and Bozkir, Efe
    and Gerjets, Peter
    and Trautwein, Ulrich
    and Kasneci, Enkelejda",
    editor="Olney, Andrew M.
    and Chounta, Irene-Angelica
    and Liu, Zitao
    and Santos, Olga C.
    and Bittencourt, Ig Ibert",
    title="Automated Assessment of Encouragement and Warmth in Classrooms Leveraging Multimodal Emotional Features and ChatGPT",
    booktitle="Artificial Intelligence in Education",
    year="2024",
    publisher="Springer Nature Switzerland",
    address="Cham",
    pages="60--74",
    isbn="978-3-031-64302-6"
}

@inproceedings{devlin-etal-2019-bert,
    title = "{BERT}: Pre-training of Deep Bidirectional Transformers for Language Understanding",
    author = "Devlin, Jacob  and
      Chang, Ming-Wei  and
      Lee, Kenton  and
      Toutanova, Kristina",
    editor = "Burstein, Jill  and
      Doran, Christy  and
      Solorio, Thamar",
    booktitle = "Proceedings of the 2019 Conference of the North {A}merican Chapter of the Association for Computational Linguistics: Human Language Technologies, Volume 1 (Long and Short Papers)",
    month = jun,
    year = "2019",
    address = "Minneapolis, Minnesota",
    publisher = "Association for Computational Linguistics",
    url = "https://aclanthology.org/N19-1423/",
    doi = "10.18653/v1/N19-1423",
    pages = "4171--4186"
}

@misc{grattafiori2024llama3herdmodels,
      title={The Llama 3 Herd of Models}, 
      author={Aaron Grattafiori and Abhimanyu Dubey and Abhinav Jauhri and Abhinav Pandey and Abhishek Kadian and Ahmad Al-Dahle and Aiesha Letman and Akhil Mathur and Alan Schelten and Alex Vaughan and Amy Yang and Angela Fan and Anirudh Goyal and Anthony Hartshorn and Aobo Yang and Archi Mitra and Archie Sravankumar and Artem Korenev and Arthur Hinsvark and Arun Rao and Aston Zhang and Aurelien Rodriguez and Austen Gregerson and Ava Spataru and Baptiste Roziere and Bethany Biron and Binh Tang and Bobbie Chern and Charlotte Caucheteux and Chaya Nayak and Chloe Bi and Chris Marra and Chris McConnell and Christian Keller and Christophe Touret and Chunyang Wu and Corinne Wong and Cristian Canton Ferrer and Cyrus Nikolaidis and Damien Allonsius and Daniel Song and Danielle Pintz and Danny Livshits and Danny Wyatt and David Esiobu and Dhruv Choudhary and Dhruv Mahajan and Diego Garcia-Olano and Diego Perino and Dieuwke Hupkes and Egor Lakomkin and Ehab AlBadawy and Elina Lobanova and Emily Dinan and Eric Michael Smith and Filip Radenovic and Francisco Guzmán and Frank Zhang and Gabriel Synnaeve and Gabrielle Lee and Georgia Lewis Anderson and Govind Thattai and Graeme Nail and Gregoire Mialon and Guan Pang and Guillem Cucurell and Hailey Nguyen and Hannah Korevaar and Hu Xu and Hugo Touvron and Iliyan Zarov and Imanol Arrieta Ibarra and Isabel Kloumann and Ishan Misra and Ivan Evtimov and Jack Zhang and Jade Copet and Jaewon Lee and Jan Geffert and Jana Vranes and Jason Park and Jay Mahadeokar and Jeet Shah and Jelmer van der Linde and Jennifer Billock and Jenny Hong and Jenya Lee and Jeremy Fu and Jianfeng Chi and Jianyu Huang and Jiawen Liu and Jie Wang and Jiecao Yu and Joanna Bitton and Joe Spisak and Jongsoo Park and Joseph Rocca and Joshua Johnstun and Joshua Saxe and Junteng Jia and Kalyan Vasuden Alwala and Karthik Prasad and Kartikeya Upasani and Kate Plawiak and Ke Li and Kenneth Heafield and Kevin Stone and Khalid El-Arini and Krithika Iyer and Kshitiz Malik and Kuenley Chiu and Kunal Bhalla and Kushal Lakhotia and Lauren Rantala-Yeary and Laurens van der Maaten and Lawrence Chen and Liang Tan and Liz Jenkins and Louis Martin and Lovish Madaan and Lubo Malo and Lukas Blecher and Lukas Landzaat and Luke de Oliveira and Madeline Muzzi and Mahesh Pasupuleti and Mannat Singh and Manohar Paluri and Marcin Kardas and Maria Tsimpoukelli and Mathew Oldham and Mathieu Rita and Maya Pavlova and Melanie Kambadur and Mike Lewis and Min Si and Mitesh Kumar Singh and Mona Hassan and Naman Goyal and Narjes Torabi and Nikolay Bashlykov and Nikolay Bogoychev and Niladri Chatterji and Ning Zhang and Olivier Duchenne and Onur Çelebi and Patrick Alrassy and Pengchuan Zhang and Pengwei Li and Petar Vasic and Peter Weng and Prajjwal Bhargava and Pratik Dubal and Praveen Krishnan and Punit Singh Koura and Puxin Xu and Qing He and Qingxiao Dong and Ragavan Srinivasan and Raj Ganapathy and Ramon Calderer and Ricardo Silveira Cabral and Robert Stojnic and Roberta Raileanu and Rohan Maheswari and Rohit Girdhar and Rohit Patel and Romain Sauvestre and Ronnie Polidoro and Roshan Sumbaly and Ross Taylor and Ruan Silva and Rui Hou and Rui Wang and Saghar Hosseini and Sahana Chennabasappa and Sanjay Singh and Sean Bell and Seohyun Sonia Kim and Sergey Edunov and Shaoliang Nie and Sharan Narang and Sharath Raparthy and Sheng Shen and Shengye Wan and Shruti Bhosale and Shun Zhang and Simon Vandenhende and Soumya Batra and Spencer Whitman and Sten Sootla and Stephane Collot and Suchin Gururangan and Sydney Borodinsky and Tamar Herman and Tara Fowler and Tarek Sheasha and Thomas Georgiou and Thomas Scialom and Tobias Speckbacher and Todor Mihaylov and Tong Xiao and Ujjwal Karn and Vedanuj Goswami and Vibhor Gupta and Vignesh Ramanathan and Viktor Kerkez and Vincent Gonguet and Virginie Do and Vish Vogeti and Vítor Albiero and Vladan Petrovic and Weiwei Chu and Wenhan Xiong and Wenyin Fu and Whitney Meers and Xavier Martinet and Xiaodong Wang and Xiaofang Wang and Xiaoqing Ellen Tan and Xide Xia and Xinfeng Xie and Xuchao Jia and Xuewei Wang and Yaelle Goldschlag and Yashesh Gaur and Yasmine Babaei and Yi Wen and Yiwen Song and Yuchen Zhang and Yue Li and Yuning Mao and Zacharie Delpierre Coudert and Zheng Yan and Zhengxing Chen and Zoe Papakipos and Aaditya Singh and Aayushi Srivastava and Abha Jain and Adam Kelsey and Adam Shajnfeld and Adithya Gangidi and Adolfo Victoria and Ahuva Goldstand and Ajay Menon and Ajay Sharma and Alex Boesenberg and Alexei Baevski and Allie Feinstein and Amanda Kallet and Amit Sangani and Amos Teo and Anam Yunus and Andrei Lupu and Andres Alvarado and Andrew Caples and Andrew Gu and Andrew Ho and Andrew Poulton and Andrew Ryan and Ankit Ramchandani and Annie Dong and Annie Franco and Anuj Goyal and Aparajita Saraf and Arkabandhu Chowdhury and Ashley Gabriel and Ashwin Bharambe and Assaf Eisenman and Azadeh Yazdan and Beau James and Ben Maurer and Benjamin Leonhardi and Bernie Huang and Beth Loyd and Beto De Paola and Bhargavi Paranjape and Bing Liu and Bo Wu and Boyu Ni and Braden Hancock and Bram Wasti and Brandon Spence and Brani Stojkovic and Brian Gamido and Britt Montalvo and Carl Parker and Carly Burton and Catalina Mejia and Ce Liu and Changhan Wang and Changkyu Kim and Chao Zhou and Chester Hu and Ching-Hsiang Chu and Chris Cai and Chris Tindal and Christoph Feichtenhofer and Cynthia Gao and Damon Civin and Dana Beaty and Daniel Kreymer and Daniel Li and David Adkins and David Xu and Davide Testuggine and Delia David and Devi Parikh and Diana Liskovich and Didem Foss and Dingkang Wang and Duc Le and Dustin Holland and Edward Dowling and Eissa Jamil and Elaine Montgomery and Eleonora Presani and Emily Hahn and Emily Wood and Eric-Tuan Le and Erik Brinkman and Esteban Arcaute and Evan Dunbar and Evan Smothers and Fei Sun and Felix Kreuk and Feng Tian and Filippos Kokkinos and Firat Ozgenel and Francesco Caggioni and Frank Kanayet and Frank Seide and Gabriela Medina Florez and Gabriella Schwarz and Gada Badeer and Georgia Swee and Gil Halpern and Grant Herman and Grigory Sizov and Guangyi and Zhang and Guna Lakshminarayanan and Hakan Inan and Hamid Shojanazeri and Han Zou and Hannah Wang and Hanwen Zha and Haroun Habeeb and Harrison Rudolph and Helen Suk and Henry Aspegren and Hunter Goldman and Hongyuan Zhan and Ibrahim Damlaj and Igor Molybog and Igor Tufanov and Ilias Leontiadis and Irina-Elena Veliche and Itai Gat and Jake Weissman and James Geboski and James Kohli and Janice Lam and Japhet Asher and Jean-Baptiste Gaya and Jeff Marcus and Jeff Tang and Jennifer Chan and Jenny Zhen and Jeremy Reizenstein and Jeremy Teboul and Jessica Zhong and Jian Jin and Jingyi Yang and Joe Cummings and Jon Carvill and Jon Shepard and Jonathan McPhie and Jonathan Torres and Josh Ginsburg and Junjie Wang and Kai Wu and Kam Hou U and Karan Saxena and Kartikay Khandelwal and Katayoun Zand and Kathy Matosich and Kaushik Veeraraghavan and Kelly Michelena and Keqian Li and Kiran Jagadeesh and Kun Huang and Kunal Chawla and Kyle Huang and Lailin Chen and Lakshya Garg and Lavender A and Leandro Silva and Lee Bell and Lei Zhang and Liangpeng Guo and Licheng Yu and Liron Moshkovich and Luca Wehrstedt and Madian Khabsa and Manav Avalani and Manish Bhatt and Martynas Mankus and Matan Hasson and Matthew Lennie and Matthias Reso and Maxim Groshev and Maxim Naumov and Maya Lathi and Meghan Keneally and Miao Liu and Michael L. Seltzer and Michal Valko and Michelle Restrepo and Mihir Patel and Mik Vyatskov and Mikayel Samvelyan and Mike Clark and Mike Macey and Mike Wang and Miquel Jubert Hermoso and Mo Metanat and Mohammad Rastegari and Munish Bansal and Nandhini Santhanam and Natascha Parks and Natasha White and Navyata Bawa and Nayan Singhal and Nick Egebo and Nicolas Usunier and Nikhil Mehta and Nikolay Pavlovich Laptev and Ning Dong and Norman Cheng and Oleg Chernoguz and Olivia Hart and Omkar Salpekar and Ozlem Kalinli and Parkin Kent and Parth Parekh and Paul Saab and Pavan Balaji and Pedro Rittner and Philip Bontrager and Pierre Roux and Piotr Dollar and Polina Zvyagina and Prashant Ratanchandani and Pritish Yuvraj and Qian Liang and Rachad Alao and Rachel Rodriguez and Rafi Ayub and Raghotham Murthy and Raghu Nayani and Rahul Mitra and Rangaprabhu Parthasarathy and Raymond Li and Rebekkah Hogan and Robin Battey and Rocky Wang and Russ Howes and Ruty Rinott and Sachin Mehta and Sachin Siby and Sai Jayesh Bondu and Samyak Datta and Sara Chugh and Sara Hunt and Sargun Dhillon and Sasha Sidorov and Satadru Pan and Saurabh Mahajan and Saurabh Verma and Seiji Yamamoto and Sharadh Ramaswamy and Shaun Lindsay and Shaun Lindsay and Sheng Feng and Shenghao Lin and Shengxin Cindy Zha and Shishir Patil and Shiva Shankar and Shuqiang Zhang and Shuqiang Zhang and Sinong Wang and Sneha Agarwal and Soji Sajuyigbe and Soumith Chintala and Stephanie Max and Stephen Chen and Steve Kehoe and Steve Satterfield and Sudarshan Govindaprasad and Sumit Gupta and Summer Deng and Sungmin Cho and Sunny Virk and Suraj Subramanian and Sy Choudhury and Sydney Goldman and Tal Remez and Tamar Glaser and Tamara Best and Thilo Koehler and Thomas Robinson and Tianhe Li and Tianjun Zhang and Tim Matthews and Timothy Chou and Tzook Shaked and Varun Vontimitta and Victoria Ajayi and Victoria Montanez and Vijai Mohan and Vinay Satish Kumar and Vishal Mangla and Vlad Ionescu and Vlad Poenaru and Vlad Tiberiu Mihailescu and Vladimir Ivanov and Wei Li and Wenchen Wang and Wenwen Jiang and Wes Bouaziz and Will Constable and Xiaocheng Tang and Xiaojian Wu and Xiaolan Wang and Xilun Wu and Xinbo Gao and Yaniv Kleinman and Yanjun Chen and Ye Hu and Ye Jia and Ye Qi and Yenda Li and Yilin Zhang and Ying Zhang and Yossi Adi and Youngjin Nam and Yu and Wang and Yu Zhao and Yuchen Hao and Yundi Qian and Yunlu Li and Yuzi He and Zach Rait and Zachary DeVito and Zef Rosnbrick and Zhaoduo Wen and Zhenyu Yang and Zhiwei Zhao and Zhiyu Ma},
      year={2024},
      eprint={2407.21783},
      archivePrefix={arXiv},
      primaryClass={cs.AI},
      url={https://arxiv.org/abs/2407.21783}, 
}

@misc{jiang2024mixtralexperts,
      title={Mixtral of Experts}, 
      author={Albert Q. Jiang and Alexandre Sablayrolles and Antoine Roux and Arthur Mensch and Blanche Savary and Chris Bamford and Devendra Singh Chaplot and Diego de las Casas and Emma Bou Hanna and Florian Bressand and Gianna Lengyel and Guillaume Bour and Guillaume Lample and Lélio Renard Lavaud and Lucile Saulnier and Marie-Anne Lachaux and Pierre Stock and Sandeep Subramanian and Sophia Yang and Szymon Antoniak and Teven Le Scao and Théophile Gervet and Thibaut Lavril and Thomas Wang and Timothée Lacroix and William El Sayed},
      year={2024},
      eprint={2401.04088},
      archivePrefix={arXiv},
      primaryClass={cs.LG},
      url={https://arxiv.org/abs/2401.04088}, 
}

@inproceedings{lundberg2017aunifiedapproach,
    author = {Lundberg, Scott M. and Lee, Su-In},
    title = {A unified approach to interpreting model predictions},
    year = {2017},
    isbn = {9781510860964},
    publisher = {Curran Associates Inc.},
    address = {Red Hook, NY, USA},
    booktitle = {Proceedings of the 31st International Conference on Neural Information Processing Systems},
    pages = {4768–4777},
    numpages = {10},
    location = {Long Beach, California, USA},
    series = {NIPS'17}
}

@book{Pianta2008ClassroomAS,
  title={Classroom Assessment Scoring System{\texttrademark}: Manual K-3.},
  author={Robert C. Pianta and Karen M. La Paro and Bridget K. Hamre},
  year={2008},
  url={https://api.semanticscholar.org/CorpusID:69768532},
  publisher={Brookes Publishing}
}

@inproceedings{demszky-hill-2023-ncte,
    title = "The {NCTE} Transcripts: A Dataset of Elementary Math Classroom Transcripts",
    author = "Demszky, Dorottya  and
      Hill, Heather",
    editor = {Kochmar, Ekaterina  and
      Burstein, Jill  and
      Horbach, Andrea  and
      Laarmann-Quante, Ronja  and
      Madnani, Nitin  and
      Tack, Ana{\"i}s  and
      Yaneva, Victoria  and
      Yuan, Zheng  and
      Zesch, Torsten},
    booktitle = "Proceedings of the 18th Workshop on Innovative Use of NLP for Building Educational Applications (BEA 2023)",
    month = jul,
    year = "2023",
    address = "Toronto, Canada",
    publisher = "Association for Computational Linguistics",
    url = "https://aclanthology.org/2023.bea-1.44/",
    doi = "10.18653/v1/2023.bea-1.44",
    pages = "528--538"
}

@article{hattie2007thepoweroffeedback,
 ISSN = {00346543, 19351046},
 URL = {http://www.jstor.org/stable/4624888},
 author = {John Hattie and Helen Timperley},
 journal = {Review of Educational Research},
 number = {1},
 pages = {81--112},
 publisher = {[Sage Publications, Inc., American Educational Research Association]},
 title = {The Power of Feedback},
 urldate = {2025-12-01},
 volume = {77},
 year = {2007}
}

@misc{jiang2023mistral7b,
      title={Mistral 7B}, 
      author={Albert Q. Jiang and Alexandre Sablayrolles and Arthur Mensch and Chris Bamford and Devendra Singh Chaplot and Diego de las Casas and Florian Bressand and Gianna Lengyel and Guillaume Lample and Lucile Saulnier and Lélio Renard Lavaud and Marie-Anne Lachaux and Pierre Stock and Teven Le Scao and Thibaut Lavril and Thomas Wang and Timothée Lacroix and William El Sayed},
      year={2023},
      eprint={2310.06825},
      archivePrefix={arXiv},
      primaryClass={cs.CL},
      url={https://arxiv.org/abs/2310.06825}, 
}

@misc{yang2025qwen3technicalreport,
      title={Qwen3 Technical Report}, 
      author={An Yang and Anfeng Li and Baosong Yang and Beichen Zhang and Binyuan Hui and Bo Zheng and Bowen Yu and Chang Gao and Chengen Huang and Chenxu Lv and Chujie Zheng and Dayiheng Liu and Fan Zhou and Fei Huang and Feng Hu and Hao Ge and Haoran Wei and Huan Lin and Jialong Tang and Jian Yang and Jianhong Tu and Jianwei Zhang and Jianxin Yang and Jiaxi Yang and Jing Zhou and Jingren Zhou and Junyang Lin and Kai Dang and Keqin Bao and Kexin Yang and Le Yu and Lianghao Deng and Mei Li and Mingfeng Xue and Mingze Li and Pei Zhang and Peng Wang and Qin Zhu and Rui Men and Ruize Gao and Shixuan Liu and Shuang Luo and Tianhao Li and Tianyi Tang and Wenbiao Yin and Xingzhang Ren and Xinyu Wang and Xinyu Zhang and Xuancheng Ren and Yang Fan and Yang Su and Yichang Zhang and Yinger Zhang and Yu Wan and Yuqiong Liu and Zekun Wang and Zeyu Cui and Zhenru Zhang and Zhipeng Zhou and Zihan Qiu},
      year={2025},
      eprint={2505.09388},
      archivePrefix={arXiv},
      primaryClass={cs.CL},
      url={https://arxiv.org/abs/2505.09388}, 
}

@misc{lan2020albertlitebertselfsupervised,
      title={ALBERT: A Lite BERT for Self-supervised Learning of Language Representations}, 
      author={Zhenzhong Lan and Mingda Chen and Sebastian Goodman and Kevin Gimpel and Piyush Sharma and Radu Soricut},
      year={2020},
      eprint={1909.11942},
      archivePrefix={arXiv},
      primaryClass={cs.CL},
      url={https://arxiv.org/abs/1909.11942}, 
}

@misc{liu2019robertarobustlyoptimizedbert,
      title={RoBERTa: A Robustly Optimized BERT Pretraining Approach}, 
      author={Yinhan Liu and Myle Ott and Naman Goyal and Jingfei Du and Mandar Joshi and Danqi Chen and Omer Levy and Mike Lewis and Luke Zettlemoyer and Veselin Stoyanov},
      year={2019},
      eprint={1907.11692},
      archivePrefix={arXiv},
      primaryClass={cs.CL},
      url={https://arxiv.org/abs/1907.11692}, 
}

@misc{he2023debertav3improvingdebertausing,
      title={DeBERTaV3: Improving DeBERTa using ELECTRA-Style Pre-Training with Gradient-Disentangled Embedding Sharing}, 
      author={Pengcheng He and Jianfeng Gao and Weizhu Chen},
      year={2023},
      eprint={2111.09543},
      archivePrefix={arXiv},
      primaryClass={cs.CL},
      url={https://arxiv.org/abs/2111.09543}, 
}

@inproceedings{wang-demszky-2023-chatgpt,
    title = "Is {C}hat{GPT} a Good Teacher Coach? Measuring Zero-Shot Performance For Scoring and Providing Actionable Insights on Classroom Instruction",
    author = "Wang, Rose  and
      Demszky, Dorottya",
    editor = {Kochmar, Ekaterina  and
      Burstein, Jill  and
      Horbach, Andrea  and
      Laarmann-Quante, Ronja  and
      Madnani, Nitin  and
      Tack, Ana{\"i}s  and
      Yaneva, Victoria  and
      Yuan, Zheng  and
      Zesch, Torsten},
    booktitle = "Proceedings of the 18th Workshop on Innovative Use of NLP for Building Educational Applications (BEA 2023)",
    month = jul,
    year = "2023",
    address = "Toronto, Canada",
    publisher = "Association for Computational Linguistics",
    url = "https://aclanthology.org/2023.bea-1.53/",
    doi = "10.18653/v1/2023.bea-1.53",
    pages = "626--667"
}

@inproceedings{sessler2025can,
  title={Can AI grade your essays? A comparative analysis of large language models and teacher ratings in multidimensional essay scoring},
  author={Se{\ss}ler, Kathrin and F{\"u}rstenberg, Maurice and B{\"u}hler, Babette and Kasneci, Enkelejda},
  booktitle={Proceedings of the 15th International Learning Analytics and Knowledge Conference},
  pages={462--472},
  year={2025}
}

@InProceedings{tran2023utilizingnatural,
author="Tran, Nhat
and Pierce, Benjamin
and Litman, Diane
and Correnti, Richard
and Matsumura, Lindsay Clare",
editor="Wang, Ning
and Rebolledo-Mendez, Genaro
and Dimitrova, Vania
and Matsuda, Noboru
and Santos, Olga C.",
title="Utilizing Natural Language Processing for Automated Assessment of Classroom Discussion",
booktitle="Artificial Intelligence in Education. Posters and Late Breaking Results, Workshops and Tutorials, Industry and Innovation Tracks, Practitioners, Doctoral Consortium and Blue Sky",
year="2023",
publisher="Springer Nature Switzerland",
address="Cham",
pages="490--496",
isbn="978-3-031-36336-8"
}

@article{MEYER2024100199,
title = {Using LLMs to bring evidence-based feedback into the classroom: AI-generated feedback increases secondary students’ text revision, motivation, and positive emotions},
journal = {Computers and Education: Artificial Intelligence},
volume = {6},
pages = {100199},
year = {2024},
issn = {2666-920X},
doi = {https://doi.org/10.1016/j.caeai.2023.100199},
url = {https://www.sciencedirect.com/science/article/pii/S2666920X23000784},
author = {Jennifer Meyer and Thorben Jansen and Ronja Schiller and Lucas W. Liebenow and Marlene Steinbach and Andrea Horbach and Johanna Fleckenstein}
}

@misc{mohammadi_explaining_2024,
    title = {Explaining {Large} {Language} {Models} {Decisions} {Using} {Shapley} {Values}},
    url = {http://arxiv.org/abs/2404.01332},
    doi = {10.48550/arXiv.2404.01332},
    urldate = {2025-10-30},
    publisher = {arXiv},
    author = {Mohammadi, Behnam},
    month = nov,
    year = {2024},
    note = {arXiv:2404.01332 [cs]},
}

@misc{ribeiro2016whyitrustyou,
      title={"Why Should I Trust You?": Explaining the Predictions of Any Classifier}, 
      author={Marco Tulio Ribeiro and Sameer Singh and Carlos Guestrin},
      year={2016},
      eprint={1602.04938},
      archivePrefix={arXiv},
      primaryClass={cs.LG},
      url={https://arxiv.org/abs/1602.04938}, 
}

@inproceedings{jain-wallace-2019-attention,
    title = "{A}ttention is not {E}xplanation",
    author = "Jain, Sarthak  and
      Wallace, Byron C.",
    editor = "Burstein, Jill  and
      Doran, Christy  and
      Solorio, Thamar",
    booktitle = "Proceedings of the 2019 Conference of the North {A}merican Chapter of the Association for Computational Linguistics: Human Language Technologies, Volume 1 (Long and Short Papers)",
    month = jun,
    year = "2019",
    address = "Minneapolis, Minnesota",
    publisher = "Association for Computational Linguistics",
    url = "https://aclanthology.org/N19-1357/",
    doi = "10.18653/v1/N19-1357",
    pages = "3543--3556"
}

@inproceedings{wiegreffe-pinter-2019-attention,
    title = "Attention is not not Explanation",
    author = "Wiegreffe, Sarah  and
      Pinter, Yuval",
    editor = "Inui, Kentaro  and
      Jiang, Jing  and
      Ng, Vincent  and
      Wan, Xiaojun",
    booktitle = "Proceedings of the 2019 Conference on Empirical Methods in Natural Language Processing and the 9th International Joint Conference on Natural Language Processing (EMNLP-IJCNLP)",
    month = nov,
    year = "2019",
    address = "Hong Kong, China",
    publisher = "Association for Computational Linguistics",
    url = "https://aclanthology.org/D19-1002/",
    doi = "10.18653/v1/D19-1002",
    pages = "11--20"
}

@inproceedings{turpin2023languagemodels,
author = {Turpin, Miles and Michael, Julian and Perez, Ethan and Bowman, Samuel R.},
title = {Language models don't always say what they think: unfaithful explanations in chain-of-thought prompting},
year = {2023},
publisher = {Curran Associates Inc.},
address = {Red Hook, NY, USA},
booktitle = {Proceedings of the 37th International Conference on Neural Information Processing Systems},
articleno = {3275},
numpages = {14},
location = {New Orleans, LA, USA},
series = {NIPS '23}
}

@inproceedings{ye2022theunreliability,
author = {Ye, Xi and Durrett, Greg},
title = {The unreliability of explanations in few-shot prompting for textual reasoning},
year = {2022},
isbn = {9781713871088},
publisher = {Curran Associates Inc.},
address = {Red Hook, NY, USA},
booktitle = {Proceedings of the 36th International Conference on Neural Information Processing Systems},
articleno = {2202},
numpages = {15},
location = {New Orleans, LA, USA},
series = {NIPS '22}
}

@inproceedings{ayala-bechard-2024-reducing,
    title = "Reducing hallucination in structured outputs via Retrieval-Augmented Generation",
    author = "Ayala, Orlando  and
      Bechard, Patrice",
    editor = "Yang, Yi  and
      Davani, Aida  and
      Sil, Avi  and
      Kumar, Anoop",
    booktitle = "Proceedings of the 2024 Conference of the North American Chapter of the Association for Computational Linguistics: Human Language Technologies (Volume 6: Industry Track)",
    month = jun,
    year = "2024",
    address = "Mexico City, Mexico",
    publisher = "Association for Computational Linguistics",
    url = "https://aclanthology.org/2024.naacl-industry.19/",
    doi = "10.18653/v1/2024.naacl-industry.19",
    pages = "228--238"
}

@inproceedings{deyoung-etal-2020-eraser,
    title = "{ERASER}: {A} Benchmark to Evaluate Rationalized {NLP} Models",
    author = "DeYoung, Jay  and
      Jain, Sarthak  and
      Rajani, Nazneen Fatema  and
      Lehman, Eric  and
      Xiong, Caiming  and
      Socher, Richard  and
      Wallace, Byron C.",
    editor = "Jurafsky, Dan  and
      Chai, Joyce  and
      Schluter, Natalie  and
      Tetreault, Joel",
    booktitle = "Proceedings of the 58th Annual Meeting of the Association for Computational Linguistics",
    month = jul,
    year = "2020",
    address = "Online",
    publisher = "Association for Computational Linguistics",
    url = "https://aclanthology.org/2020.acl-main.408/",
    doi = "10.18653/v1/2020.acl-main.408",
    pages = "4443--4458"
}

@inproceedings{jacovi-goldberg-2020-towards,
    title = "Towards Faithfully Interpretable {NLP} Systems: How Should We Define and Evaluate Faithfulness?",
    author = "Jacovi, Alon  and
      Goldberg, Yoav",
    editor = "Jurafsky, Dan  and
      Chai, Joyce  and
      Schluter, Natalie  and
      Tetreault, Joel",
    booktitle = "Proceedings of the 58th Annual Meeting of the Association for Computational Linguistics",
    month = jul,
    year = "2020",
    address = "Online",
    publisher = "Association for Computational Linguistics",
    url = "https://aclanthology.org/2020.acl-main.386/",
    doi = "10.18653/v1/2020.acl-main.386",
    pages = "4198--4205"
}

@INPROCEEDINGS{khediri2024enhancingmachine,
  author={Khediri, Abderrazak and Slimi, Hamda and Yahiaoui, Ayoub and Derdour, Makhlouf and Bendjenna, Hakim and Ghenai, Charaf Eddine},
  booktitle={2024 6th International Conference on Pattern Analysis and Intelligent Systems (PAIS)}, 
  title={Enhancing Machine Learning Model Interpretability in Intrusion Detection Systems through SHAP Explanations and LLM-Generated Descriptions}, 
  year={2024},
  volume={},
  number={},
  pages={1-6},
  doi={10.1109/PAIS62114.2024.10541168}}

@misc{eu_ai_act_2024,
  title        = {Regulation (EU) 2024/1689 of the European Parliament and of the Council of 13 June 2024 Laying Down Harmonised Rules on Artificial Intelligence (Artificial Intelligence Act)},
  author       = {{European Parliament and Council of the European Union}},
  year         = {2024},
  howpublished = {Official Journal of the European Union, OJ L, 2024/1689, 12.7.2024},
  note         = {Entered into force on 1 August 2024},
}

@article{FUTTERER2026102264,
title = {Validating automated assessments of teaching effectiveness using multimodal data},
journal = {Learning and Instruction},
volume = {101},
pages = {102264},
year = {2026},
issn = {0959-4752},
doi = {https://doi.org/10.1016/j.learninstruc.2025.102264},
url = {https://www.sciencedirect.com/science/article/pii/S0959475225001884},
author = {Tim Fütterer and Ruikun Hou and Babette Bühler and Efe Bozkir and Courtney Bell and Enkelejda Kasneci and Peter Gerjets and Ulrich Trautwein}
}

@inproceedings{hou-etal-2025-llm,
    title = "{LLM}-Human Alignment in Evaluating Teacher Questioning Practices: Beyond Ratings to Explanation",
    author = {Hou, Ruikun  and
      F{\"u}tterer, Tim  and
      B{\"u}hler, Babette  and
      Schreyer, Patrick  and
      Gerjets, Peter  and
      Trautwein, Ulrich  and
      Kasneci, Enkelejda},
    editor = "Wilson, Joshua  and
      Ormerod, Christopher  and
      Beiting Parrish, Magdalen",
    booktitle = "Proceedings of the Artificial Intelligence in Measurement and Education Conference (AIME-Con): Full Papers",
    month = oct,
    year = "2025",
    address = "Wyndham Grand Pittsburgh, Downtown, Pittsburgh, Pennsylvania, United States",
    publisher = "National Council on Measurement in Education (NCME)",
    url = "https://aclanthology.org/2025.aimecon-main.26/",
    pages = "239--249",
    ISBN = "979-8-218-84228-4"
}

@article{barocas2016big,
  title={Big Data's Disparate Impact},
  author={Barocas, Solon and Selbst, Andrew D.},
  journal={California Law Review},
  year={2016},
  doi={https://doi.org/10.15779/Z38BG31}
}

@article{mehrabi2021survey,
author = {Mehrabi, Ninareh and Morstatter, Fred and Saxena, Nripsuta and Lerman, Kristina and Galstyan, Aram},
title = {A Survey on Bias and Fairness in Machine Learning},
year = {2021},
issue_date = {July 2022},
publisher = {Association for Computing Machinery},
address = {New York, NY, USA},
volume = {54},
number = {6},
issn = {0360-0300},
url = {https://doi.org/10.1145/3457607},
doi = {10.1145/3457607},
journal = {ACM Comput. Surv.},
month = jul,
articleno = {115},
numpages = {35}
}

@article{holmes2022ethics,
  title   = {Ethics of {AI} in Education: Towards a Community-Wide Framework},
  author  = {Holmes, Wayne and
             Porayska-Pomsta, Kaska and
             Holstein, Ken and
             Sutherland, Emma and
             Baker, Toby and
             Shum, Simon Buckingham and
             Santos, Olga C. and
             Rodrigo, Mercedes T. and
             Cukurova, Mutlu and
             Bittencourt, Ig Ibert and
             Koedinger, Kenneth R.},
  journal = {International Journal of Artificial Intelligence in Education},
  year    = {2022},
  volume  = {32},
  number  = {3},
  pages   = {504--526},
  doi     = {10.1007/s40593-021-00239-1}
}

@article{nguyen2023ethical,
  title   = {Ethical Principles for Artificial Intelligence in Education},
  author  = {Nguyen, Andy and
             Ngo, Ha Ngan and
             Hong, Yvonne and
             Dang, Belle and
             Nguyen, Bich-Phuong Thi},
  journal = {Education and Information Technologies},
  year    = {2023},
  volume  = {28},
  number  = {4},
  pages   = {4221--4241},
  doi     = {10.1007/s10639-022-11316-w}
}

@InProceedings{boulanger2020shaped,
author="Boulanger, David
and Kumar, Vivekanandan",
editor="Kumar, Vivekanandan
and Troussas, Christos",
title="SHAPed Automated Essay Scoring: Explaining Writing Features' Contributions to English Writing Organization",
booktitle="Intelligent Tutoring Systems",
year="2020",
publisher="Springer International Publishing",
address="Cham",
pages="68--78",
isbn="978-3-030-49663-0"
}

@ARTICLE{kumar2020explainable,
    
AUTHOR={Kumar, Vivekanandan  and Boulanger, David },
           
TITLE={Explainable Automated Essay Scoring: Deep Learning Really Has Pedagogical Value},
          
JOURNAL={Frontiers in Education},
          
VOLUME={Volume 5 - 2020},
  
YEAR={2020},
  
URL={https://www.frontiersin.org/journals/education/articles/10.3389/feduc.2020.572367},
  
DOI={10.3389/feduc.2020.572367},
  
ISSN={2504-284X}}

\appendix
\section{Training Hyperparameters}
\label{app:hyperparams}

The model was fine-tuned using the Hugging Face \texttt{Trainer} API with the following key hyperparameters:

\begin{itemize}
    \item \textbf{Learning rate:} $1 \times 10^{-5}$
    \item \textbf{Batch size:} 1 per device (with gradient accumulation of 8 steps)
    \item \textbf{Number of epochs:} 10
    \item \textbf{Weight decay:} 0.01
    \item \textbf{Maximum gradient norm:} 1.0
    \item \textbf{Evaluation strategy:} per epoch
    \item \textbf{Saving strategy:} per epoch (keeping best model based on validation MAE)
    \item \textbf{Learning rate scheduler:} cosine schedule
    \item \textbf{Warmup ratio:} 0.1
    \item \textbf{Mixed precision:} FP16 when available
    \item \textbf{Early stopping:} patience of 3 validation checks
\end{itemize}

All other hyperparameters were kept at their default values. One single H200 GPU was used and computations took a total of $\sim$322 hours ($\sim$20 hours for PLMs and $\sim$302 hours for LLMs). All experiments could have been done in a single A100 GPU, since we used the quantized versions of the models, except for the LLMs Qwen3~235B~A22B~Instruct, and Mixtral~8$\times$22B~Instruct, which require more than 80GB VRAM, even in their 4-bit \texttt{nf4} quantized versions.

\section{Prompts}
\label{app:prompts}
Fig.~\ref{fig:llmscoringprompt} shows an example of the few-shot prompt transcript Quality of Feedback scoring, and Fig.~\ref{fig:llmrankingpromp} show the zero-shot prompt for sentence ranking.

\begin{figure*}
    \centering
    \begin{tcolorbox}[colback=back, colframe=frame, boxrule=3pt, left=0.5em, right=0.5em, top=0.5em, bottom=0.5em,]
    % \fontsize{8}{9}\selectfont
    You are a rater for classroom transcripts. You are tasked with evaluating the transcripts to rate the dimension 'Quality of Feedback' as one of the dimensions of the domain 'Instructional Support' according to the Classroom Assessment Scoring System (CLASS). 
    
    Your goal is to rate the provided transcript by focusing on the teacher's interactions labeled as "Teacher" and the students' responses labeled as "Student".

    \vspace{10pt}
    
    **Your Task:**
    
    Please read the following classroom transcript and assign a rating based on the Quality of Feedback dimension. 
    
    Quality of Feedback assesses the degree to which feedback expands and extends learning and understanding and encourages student participation. In upper elementary classrooms, significant feedback may also be provided by peers. Regardless of the source, the focus here should be on the nature of the feedback provided and the extent to which it “pushes” learning.

    \vspace{10pt}
    
    Here are examples for the different score values in the Quality of Feedback dimension:
    
    An Example for score value 1 is \{\textit{Transcript segment from training set with score 1}\}
    
    An Example for score value 2 is \{\textit{Transcript segment from training set with score 2}\}
    
    An Example for score value 3 is \{\textit{Transcript segment from training set with score 3}\}
    
    An Example for score value 4 is \{\textit{Transcript segment from training set with score 4}\}
    
    An Example for score value 5 is \{\textit{Transcript segment from training set with score 5}\}
    
    An Example for score value 6 is \{\textit{Transcript segment from training set with score 6}\}
    
    An Example for score value 7 is \{\textit{Transcript segment from training set with score 7}\}

    \vspace{10pt}
    
    You must begin your answer with
    
    `\#\#\# Rating: <the score on a 1-7 integer scale here>'

    \vspace{10pt}
    
    \#\#\# Transcript:
    
    \hspace{10pt} \{\textit{Transcript segment to score}\}

    \vspace{10pt}
    
    \#\#\# Rating:
    \end{tcolorbox}

    \caption{Prompts used for few-shot prompting of LLMs for the task of Quality of Feedback scoring of a classroom transcript segment. \{\textit{Transcript segment from training set with score 1-7}\} represents an actual example from the training set, where the specific score was given to the transcript by an expert annotator, and \{\textit{Transcript segment to score}\} represents the transcript segment to be scored.}
    \label{fig:llmscoringprompt}
\end{figure*}

\begin{figure*}
    \centering
    \begin{tcolorbox}[colback=back, colframe=frame, boxrule=3pt, left=0.5em, right=0.5em, top=0.5em, bottom=0.5em,]
    %\fontsize{10}{9}\selectfont
    You are an expert in classroom discourse analysis and the CLASS (Classroom Assessment Scoring System) framework. You will receive:

    \vspace{10pt}

    1. A transcript of a classroom interaction.
    
    2. The same transcript divided by numbered sentences, formatted as:
    
    \hspace{10pt} (1) - <SENTENCE \#1>
    
    \hspace{10pt} (2) - <SENTENCE \#2>
       
    \hspace{10pt} ...
       
    3. A score (1-7) for the "Quality of Feedback" domain according to the CLASS framework, annotated by an expert. 1 indicates very low quality feedback, while 7 indicates very high quality feedback.

    \vspace{10pt}
    
    Your task:
    
    * Identify the sentences (by their numbers) that most strongly influenced this score.
    
    * Focus specifically on aspects relevant to the "Quality of Feedback" domain (e.g., scaffolding, prompting, encouragement of thought processes, questioning strategies, or absence of these features).
    
    * Return exactly 10 sentence numbers, ordered by estimated importance (most influential first).
    
    * If fewer than 10 sentences exist, return only those available.

    \vspace{10pt}
    
    Format your response **strictly** as a single line of comma-separated numbers (no spaces, no brackets, no explanations). For example: 9,15,6,37,2,7,8,1,64,66

    \vspace{10pt}
    
    Do not include any commentary, reasoning, or additional text.

    \vspace{10pt}
    
    You must begin your answer with
    
    `\#\#\# Output: <your comma-separated list>'

    \vspace{10pt}

    1. Transcript:
    
    \hspace{10pt} \{Full transcript segment\}

    \vspace{10pt}
    
    2. Numbered Sentences:
    
    \hspace{10pt} (1) - \{First sentence\}
    
    \hspace{10pt} (2) - \{Second sentence\}
    
    \hspace{10pt} (3) - \{Third sentence\}

    \hspace{10pt} ...

    \vspace{10pt}
    
    3. Score: 
    
    \hspace{10pt} \{Previously predicted score\}

    \vspace{10pt}
    
    \#\#\# Output:
    \end{tcolorbox}

    \caption{Zero-shot prompt for LLM sentence ranking. \{Full transcript segment\} represents the full transcript segment, with the same formatting as it appears in the dataset, \{First/Second/Third/... sentence\} represents the transcript split into sentences and numbered by order of appearance, and \{Previously predicted score\} represents the Quality of Feedback score predicted for the transcript segment either by the LLMs or by the PLMs.}
    \label{fig:llmrankingpromp}
\end{figure*}

\section{Prediction Statistics}
\label{app:pred_stats}

Table~\ref{tab:prediction_distribution} presents the full prediction statistics for PLMs and LLMs. It shows the mean values for prediction, standard deviation, and if any label was not present in the predictions of a specific model.

\begin{table}[!ht]
    \centering
    \scalebox{0.7}{
    \begin{tabular}{l c c l}
        \hline
        \textbf{Model} & \textbf{Mean} & \textbf{Std. Dev.} & \textbf{Labels Missing} \\
        \hline
        Data (Training and Test)       & 4.22 & 1.14 & -- \\
        \hline
        \multicolumn{4}{l}{} \\
        \hline
        ALBERT base          & 4.10 & 0.28 & 1, 2, 6, 7 \\
        ALBERT large         & 4.33 & 0.41 & 1, 2, 7 \\
        RoBERTa base         & 4.27 & 0.22 & 1, 2, 6, 7 \\
        RoBERTa large        & 4.01 & 0.11 & 1, 2, 6, 7 \\
        BERT base            & 4.15 & 0.38 & 1, 6, 7 \\
        BERT large           & 4.14 & 0.41 & 1, 2, 7 \\
        DeBERTa V3 base      & 4.49 & 0.47 & 1, 7 \\
        DeBERTa V3 large     & 4.14 & 0.16 & 1, 2, 6, 7 \\
        \hline
        \multicolumn{4}{l}{} \\
        \hline
        Llama 3.1 8B Instruct        & 4.91 & 1.60 & -- \\
        Llama 3.1 70B Instruct       & 2.95 & 1.65 & -- \\
        Mistral Small Instruct       & 4.37 & 0.77 & 1, 7 \\
        Mistral Small 24B Instruct   & 2.55 & 0.69 & 6, 7 \\
        Mixtral 8x7B Instruct        & 3.02 & 0.52 & 6, 7 \\
        Mixtral 8x22B Instruct       & 4.30 & 1.19 & -- \\
        Qwen3 4B Instruct            & 3.96 & 1.12 & -- \\
        Qwen3 30B A3B Instruct       & 2.92 & 1.00 & 6, 7 \\
        Qwen3 235B A22B Instruct     & 2.80 & 1.13 & 7 \\
        \hline
    \end{tabular}

    }
    \caption{Prediction statistics for fine-tuned PLMs and LLMs. We report the mean and standard deviation of predicted QoF scores on the test set, as well as the CLASS labels (1--7) that were never predicted by each model.}
    \label{tab:prediction_distribution}
\end{table}

\section{Average Number of Sentences Predicted}
\label{app:number_of_sents}
Table~\ref{tab:num_of_sentences} reports the average number of sentences output by each LLM, together with the absolute difference ($\Delta$) from the expected average of 9.931 sentences.

\begin{table}[!ht]
    \centering
    \scalebox{0.8}{
    \begin{tabular}{l c c}
        \hline
        \textbf{Model} & \textbf{\# of sentences} & \textbf{$\Delta$} \\
        \hline
        Llama 3.1 8B Instruct       & 9.939 & \textbf{-0.008} \\
        Llama 3.1 70B Instruct      & 9.886 & 0.045 \\
        Mistral Small Instruct      & 9.654 & 0.277 \\
        Mistral Small 24B Instruct  & 9.871 & 0.060 \\
        Mixtral 8x7B Instruct       & 9.291 & 0.640 \\
        Mixtral 8x22B Instruct      & 9.836 & 0.095 \\
        Qwen3 4B Instruct           & 9.711 & 0.220 \\
        Qwen3 30B A3B Instruct      & 9.828 & 0.102 \\
        Qwen3 235B A22B Instruct    & 9.516 & 0.415 \\
        \hline
    \end{tabular}
    }
    \caption{Average number of sentences each model outputs for sentence ranking, and the difference between the expected and observed number of sentences.}
    \label{tab:num_of_sentences}
\end{table}

\section{Additional Details on Explanation Alignment}
\label{sec:alignment_appendix}

Table~\ref{tab:shap_llm_alignment} reports the alignment between SHAP-based and LLM-based explanations across 1,230 transcripts. 
Jaccard similarity is computed between the sets of the top-10 sentences identified by each method for every transcript.
Spearman rank correlation is computed between sentence importance rankings derived from absolute SHAP values and LLM deletion order.
Spearman correlations are reported only for transcripts where the correlation is well-defined, resulting in between 1,226 and 1,228 transcripts per model.
Multiple SHAP runs are aggregated by averaging absolute SHAP values per sentence prior to ranking.

\begin{table}[t]
    \centering
    \scalebox{0.7}{
    \begin{tabular}{lcc}
    \toprule
    \textbf{Model} & \textbf{Jaccard@10} & \textbf{Spearman $\rho$} \\
    \midrule
    Llama 3.1 8B Instruct       & $0.081 \pm 0.135$ & $0.048 \pm 0.134$ \\
    Llama 3.1 70B Instruct      & $0.089 \pm 0.136$ & $0.067 \pm 0.129$ \\
    Mistral Small Instruct      & $0.081 \pm 0.128$ & $0.054 \pm 0.129$ \\
    Mistral Small 24B Instruct  & $0.090 \pm 0.141$ & \textbf{0.082} $\pm 0.138$ \\
    Mixtral 8x7B Instruct       & $0.072 \pm 0.131$ & $0.029 \pm 0.125$ \\
    Mixtral 8x22B Instruct      & $0.088 \pm 0.142$ & $0.070 \pm 0.134$ \\
    Qwen3 4B Instruct           & $0.084 \pm 0.139$ & $0.057 \pm 0.138$ \\
    Qwen3 30B A3B Instruct      & $0.089 \pm 0.154$ & $0.071 \pm 0.134$ \\
    Qwen3 235B A22B Instruct    & \textbf{0.095} $\pm 0.159$ & $0.081 \pm 0.142$ \\
    \midrule
    \textbf{Average}            & $\mathbf{0.085}$ & $\mathbf{0.062}$ \\
    \bottomrule
    \end{tabular}
    }
    \caption{Alignment between SHAP and LLM explanations measured by Jaccard similarity over top-10 sentences and Spearman rank correlation. Values are reported as mean $\pm$ standard deviation.}\label{tab:shap_llm_alignment}
\end{table}

\section{Results of Sentence Removal}
\label{app:full_sent_removal}
\begin{figure}[!ht]
    \centering
    \includegraphics[width=\linewidth]{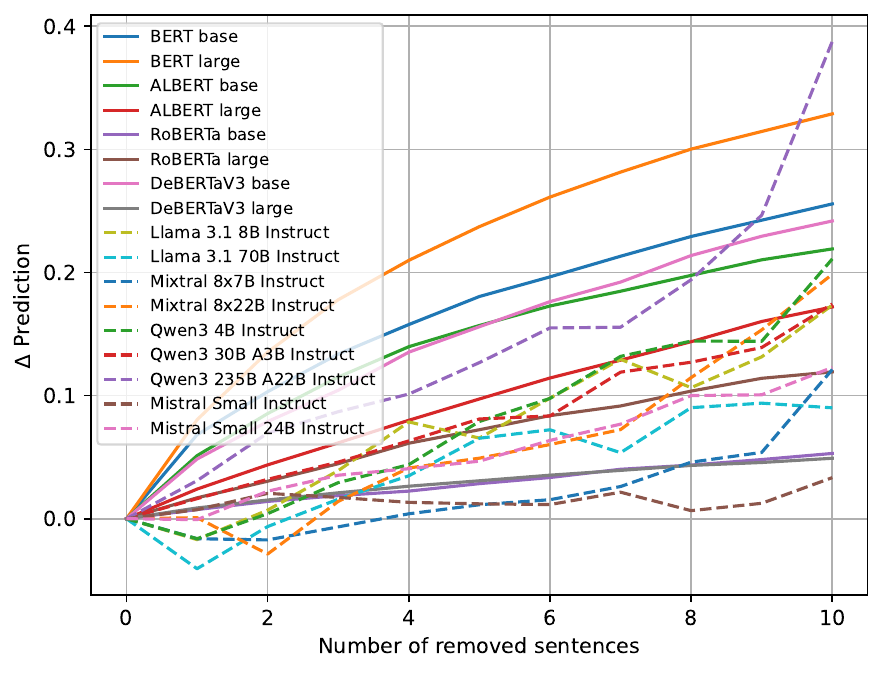}
    \caption{Average prediction changes after removing most important sentences and re-scoring within model family.}
    \label{fig:pred_changes_llm_ft}
\end{figure}

Table~\ref{tab:sentence_removal} shows the cumulative delta values after each step of the top 10 sentences removal. Fig.~\ref{fig:pred_changes_llm_ft}, represents it in a graphical form.

\begin{table*}[t]
    \centering
    \small
    \setlength{\tabcolsep}{6pt}
    \begin{tabular}{lcccccccccc}
    \toprule
    \textbf{Model} & \multicolumn{10}{c}{\textbf{Number of Sentences Removed}} \\
     & \textbf{1} & \textbf{2} & \textbf{3} & \textbf{4} & \textbf{5} & \textbf{6} & \textbf{7} & \textbf{8} & \textbf{9} & \textbf{10} \\
    \midrule
    \multicolumn{11}{c}{\textbf{PLMs}} \\
    \midrule
    ALBERT base       & 0.051 & 0.085 & 0.115 & 0.140 & 0.157 & 0.173 & 0.185 & 0.198 & 0.210 & 0.219 \\
    ALBERT large      & 0.024 & 0.044 & 0.062 & 0.080 & 0.097 & 0.114 & 0.129 & 0.144 & 0.160 & 0.172 \\
    RoBERTa base      & 0.007 & 0.014 & 0.019 & 0.022 & 0.028 & 0.033 & 0.040 & 0.044 & 0.048 & 0.053 \\
    BERT base         & 0.068 & 0.103 & 0.134 & 0.158 & 0.180 & 0.196 & 0.213 & 0.229 & 0.242 & 0.256 \\
    BERT large        & 0.080 & 0.135 & 0.178 & 0.210 & 0.237 & 0.261 & 0.282 & 0.300 & 0.315 & \textbf{0.329} \\
    DeBERTaV3 base    & 0.048 & 0.079 & 0.104 & 0.135 & 0.156 & 0.176 & 0.192 & 0.214 & 0.229 & 0.242 \\
    DeBERTaV3 large   & 0.009 & 0.015 & 0.021 & 0.026 & 0.031 & 0.035 & 0.039 & 0.043 & 0.046 & 0.049 \\
    \midrule
    \multicolumn{11}{c}{\textbf{LLMs}} \\
    \midrule
    Llama 3.1 8B Instruct      & -0.017 & 0.007 & 0.039 & 0.079 & 0.065 & 0.098 & 0.129 & 0.106 & 0.131 & 0.174 \\
    Llama 3.1 70B Instruct     & -0.041 & -0.006 & 0.016 & 0.035 & 0.065 & 0.072 & 0.054 & 0.090 & 0.094 & 0.090 \\
    Mistral Small Instruct     & 0.007 & 0.021 & 0.017 & 0.013 & 0.012 & 0.011 & 0.022 & 0.007 & 0.012 & 0.033 \\
    Mistral Small 24B Instruct & -0.001 & 0.022 & 0.035 & 0.041 & 0.047 & 0.064 & 0.077 & 0.100 & 0.101 & 0.123 \\
    Mixtral 8x7B Instruct      & -0.016 & -0.017 & -0.007 & 0.004 & 0.011 & 0.015 & 0.026 & 0.046 & 0.054 & 0.121 \\
    Mixtral 8x22B Instruct     & 0.001 & -0.029 & 0.014 & 0.041 & 0.049 & 0.060 & 0.072 & 0.115 & 0.153 & 0.199 \\
    Qwen3 4B Instruct          & -0.016 & 0.004 & 0.029 & 0.044 & 0.079 & 0.098 & 0.132 & 0.144 & 0.144 & 0.211 \\
    Qwen3 30B A3B Instruct     & 0.016 & 0.032 & 0.046 & 0.063 & 0.081 & 0.083 & 0.119 & 0.127 & 0.139 & 0.174 \\
    Qwen3 235B A22B Instruct   & 0.031 & 0.070 & 0.087 & 0.101 & 0.127 & 0.155 & 0.156 & 0.194 & 0.247 & \textbf{0.388} \\
    \bottomrule
    \end{tabular}

    \caption{Performance deltas after removing consecutive numbers of sentences. Top: PLMs. Bottom: LLMs.}
    \label{tab:sentence_removal}
\end{table*}

\section{Removed Sentence Examples}
\label{app:sent_examples}
Table~\ref{tab:removed_sentences} lists randomly selected sentences removed during the sentence-deletion experiments. Sentences are grouped by the explanation method(s) that selected them.

\begin{table*}[ht]
\centering
\small
\begin{tabular}{p{1cm} p{12cm}}
\toprule
\textbf{Model} & \textbf{Sentence} \\
\toprule \\
\multirow{10}{*}{LLM}
& - You think it's not a polygon because it has a curved side. \\
& - Teacher: Okay, what you do is, yeah, you cut it in half again so you have their equivalents. \\
& - Remember in math I spent a lot of time going over with you every time you answer a question you have to a? \\
& - Is this area right here inside the O? \\
& - What's ten times 32? \\
& - What does she need to put right next to the inches? \\
& - Student: Um - Teacher: Try it - try a number for X and see how well it works for you. \\
& - Teacher: Add my two sums of my two grids. \\
& - Technically if you're finding area, you're finding the number of squares inside the shape, correct? \\
& - So are they different? \\
\\
\cline{1-2}
\\
\multirow{10}{*}{PLM}
& - 1, 2, 3. \\
& - Teacher: I could divide it by 2, yeah. \\
& - Why don't we count on the 9th? \\
& - Tell somebody in your group what you learned in math today. \\
& - What happens to 6? \\
& - 635. \\
& - Student: Can we write an answer sentence? \\
& - Multiple Students: Label. \\
& - Student: ... \\
& - Teacher: Plus 19 - Student D, what's 2 plus 9? \\
\\
\cline{1-2}
\\
\multirow{10}{*}{Both}
& - And now I can find my total which will be what, Student H? \\
& - Student: It's going to get bigger. \\
& - Teacher: So what is the denominator there? \\
& - As you're doing this and you're answering the questions I want to label it. \\
& - So what is 12 divided by 2? \\
& - Okay so 85 plus 15 would bring me to my next whole. \\
& - What happens to 6? \\
& - Teacher: You have to go ahead .... \\
& - Multiple Students: Less. \\
& - Student: It's three times bigger. \\
\\
\bottomrule
\end{tabular}
\caption{Example sentences removed during sentence-deletion experiments, grouped by whether they were selected by LLM-based explanations, PLM-based explanations, or by both.}
\label{tab:removed_sentences}
\end{table*}

\end{document}